\documentclass[pmlr,twocolumn,10pt]{jmlr} 

\usepackage{booktabs}
\usepackage{siunitx}
\usepackage{graphicx}
\usepackage{xcolor}
\usepackage{mathtools}
\usepackage{hyperref}
\usepackage{url}
\usepackage{booktabs}
\usepackage{wrapfig}
\usepackage{multirow}
\usepackage{sidecap}

\theorembodyfont{\upshape}
\theoremheaderfont{\scshape}
\theorempostheader{:}
\theoremsep{\newline}

\jmlrvolume{LEAVE UNSET}
\jmlryear{2023}
\jmlrsubmitted{LEAVE UNSET}
\jmlrpublished{LEAVE UNSET}
\jmlrworkshop{Conference on Health, Inference, and Learning (CHIL) 2023} 

\title[Modeling Multivariate Biosignals With GNNs and Structured State Spaces]{Modeling Multivariate Biosignals With Graph Neural Networks and Structured State Space Models}

\author{%
\Name{Siyi Tang} \Email{siyitang@stanford.edu}\\
\addr Stanford University
\AND
\Name{Jared A. Dunnmon} \Email{jdunnmon@stanford.edu}\\
\addr Stanford University
\AND
\Name{Liangqiong Qu} \Email{liangqqu@hku.hk}\\
\addr University of Hong Kong
\AND
\Name{Khaled K. Saab} \Email{ksaab@stanford.edu}\\
\addr Stanford University
\AND
\Name{Tina Baykaner} \Email{tina4@stanford.edu}\\
\addr Stanford University
\AND
\Name{Christopher Lee-Messer} \Email{cleemess@stanford.edu}\\
\addr Stanford University
\AND
\Name{Daniel L. Rubin} \Email{rubin@stanford.edu}\\
\addr Stanford University
}

\begin{document}

\maketitle

\begin{abstract}
Multivariate biosignals are prevalent in many medical domains, such as electroencephalography, polysomnography, and electrocardiography. Modeling spatiotemporal dependencies in multivariate biosignals is challenging due to (1) long-range temporal dependencies and (2) complex spatial correlations between the electrodes. To address these challenges, we propose representing multivariate biosignals as time-dependent graphs and introduce \textsc{GraphS4mer}, a general graph neural network (GNN) architecture that improves performance on biosignal classification tasks by modeling spatiotemporal dependencies in biosignals. Specifically, (1) we leverage the Structured State Space architecture, a state-of-the-art deep sequence model, to capture long-range temporal dependencies in biosignals and (2) we propose a graph structure learning layer in \textsc{GraphS4mer} to learn dynamically evolving graph structures in the data.  We evaluate our proposed model on three distinct biosignal classification tasks and show that \textsc{GraphS4mer} consistently improves over existing models, including (1) seizure detection from electroencephalographic signals, outperforming a previous GNN with self-supervised pre-training by 3.1 points in AUROC; (2) sleep staging from polysomnographic signals, a 4.1 points improvement in macro-F1 score compared to existing sleep staging models; and (3) 12-lead electrocardiogram classification, outperforming previous state-of-the-art models by 2.7 points in macro-F1 score.
\end{abstract}

\paragraph*{Data and Code Availability}
All the datasets used in our study are publicly available. The Temple University Hospital EEG Seizure Corpus (TUSZ) is publicly available at \url{https://isip.piconepress.com/projects/tuh_eeg/html/downloads.shtml}. The DOD-H dataset is publicly available at \url{https://github.com/Dreem-Organization/dreem-learning-open}. The ICBEB ECG dataset is publically available at \url{http://2018.icbeb.org/Challenge.html}. Source code is publicly available at \url{https://github.com/tsy935/graphs4mer}.

\section*{Institutional Review Board (IRB)}
Since all the datasets used in this study are publicly available, this study does not require IRB approval.

\section{Introduction}
\label{sec:intro}

Multivariate biosignals are measured by multiple electrodes (sensors) and play critical roles in many medical domains. For example, electroencephalograms (EEGs), which measure brain electrical activity using sensors placed on a person's scalp, are the most common tests for seizure diagnosis. 12-lead electrocardiograms (ECGs), which assess heart electrical activity using electrodes placed on the surface of a person's body, are crucial to diagnosing heart rhythm disorders or other disease states of the heart muscle.

Several challenges exist in modeling spatiotemporal dependencies in multivariate biosignals. First, most biosignals are sampled at a high sampling rate, which results in long sequences that can be up to tens of thousands of time steps. Moreover, biosignals often involve long-range temporal correlations \citep{Berthouze2010-qf, Peng1995-mg}. Prior studies on modeling biosignals often preprocess the raw signals using frequency transformations \citep{Tang2022-gnn-ssl, Asif2020-seizurenet, Shoeibi2021-seizure-review, Covert2019-wi, Guillot2020-dreem, Guillot2021-robustsleepnet} or divide the signals into short windows and aggregate model predictions post-hoc \citep{Phan2022-sleep-review, Pradhan2022, Strodthoff2021-tx}. However, such preprocessing steps may discard important information encoded in raw signals, as well as neglect long-range temporal dependencies in the signals. Therefore, a model that is capable of modeling long-range temporal correlations is needed to better capture temporal dependencies in biosignals.

\begin{figure*}[htbp]
    \centering
    \includegraphics[width=0.85\textwidth]{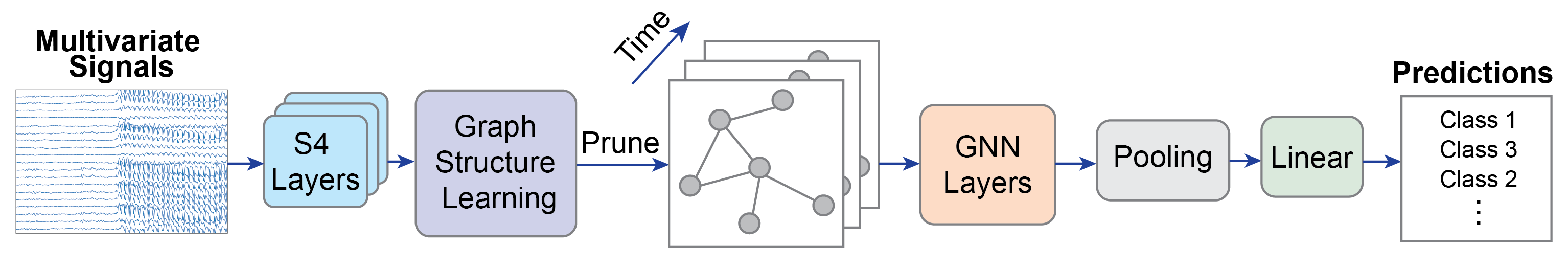}
    \caption{\textbf{Architecture of \textsc{GraphS4mer}}. The model has three main components: (1) stacked S4 layers to learn temporal dependencies in each sensor independently; (2) a graph structure learning (GSL) layer to learn dynamically evolving graph structures; (3) GNN layers to learn spatial dependencies in biosignals based on S4 embeddings and learned graph structures.}
    \label{fig:model}
\end{figure*}

Deep sequence models, including recurrent neural networks (RNNs), convolutional neural networks (CNNs), and Transformers, have specialized variants for handling long sequences \citep{Arjovsky2016-uc, erichson2021-lipschitz, Katharopoulos2020-cg, choromanski2021-rethinking}. However, they struggle to scale to long sequences of tens of thousands of time steps \citep{Tay2020-lra}. Recently, the Structured State Space sequence model (S4) \citep{Gu2022-s4}, a deep sequence model based on the classic state space approach, has outperformed previous state-of-the-art models on challenging long sequence modeling tasks, such as the Long Range Arena benchmark \citep{Tay2020-lra}, raw speech classification \citep{Gu2022-s4}, and audio generation \citep{goel22a-audio}.

Second, sensors have complex, non-Euclidean spatial correlations \citep{Garcia2013-du, Sazgar2019-bd}. For example, EEG sensors measure highly correlated yet unique electrical activity from different brain regions \citep{Michel2012-eeg}. Graphs are data structures that can represent complex, non-Euclidean correlations in the data \citep{Chami2022-graph, Bronstein2017-mj}. Previous works have adopted temporal graph neural networks (GNNs) in modeling multivariate time series, such as EEG-based seizure detection \citep{Covert2019-wi} and classification \citep{Tang2022-gnn-ssl}, traffic forecasting \citep{li2018-dcrnn, Wu2019-graph-wavenet, Zheng2020-gman, Jiang2022-survey, Tian2021-av}, and pandemic forecasting \citep{Panagopoulos2021-ng, Kapoor2020}. Nevertheless, most of these studies use sequences up to hundreds of time steps and require a predefined, static graph structure. However, the graph structure of multivariate biosignals may not be easily defined due to unknown sensor locations. For instance, while EEG sensors are typically placed according to the 10-20 standard placement \citep{Jasper1958-1020}, the exact locations of sensors vary in each individual's recordings due to the variability in individual head size, clinical need, or patient injury. Moreover, the underlying graph connectivity can evolve over time due to changes in the underlying biology. For example, brain electrical activity is more synchronized across brain areas during the event of a generalized seizure than a normal state \citep{Gloor1990-vv, Amor2009-tk}, and thus the underlying EEG graph structure can change dynamically when the EEG evolves from the normal state to a seizure state. Similarly, the heart is associated with complex dynamic signals in both normal and diseased states \citep{Qu2014-it}, and thus the underlying graph structure in ECG can change dynamically over time if a patient develops an abnormal heart rhythm. Hence, the model's ability to dynamically learn the underlying graph structures is highly desirable.

Graph structure learning (GSL) aims to jointly learn an optimized graph structure and its node and graph representations \citep{Zhu2021-gsl-survey}. GSL has been employed for spatiotemporal modeling of traffic flows \citep{Zhang2020-cl, Tang2022-traffic, shang2021-discrete, Wu2019-graph-wavenet, Bai2020-la}, irregularly sampled multivariate time series \citep{zhang2022-raindrop}, functional magnetic resonance imaging (fMRI) \citep{El-Gazzar2021, El-Gazzar2022-graphs4}, and sleep staging \citep{jia2020-graphsleepnet}. However, these studies are limited to sequences of less than 1k time steps and do not capture dynamic graphs evolving over time.

In this study, we address the foregoing challenges by (1) leveraging S4 to enable long-range temporal modeling in biosignals, (2) proposing a graph structure learning layer to learn dynamically evolving graph structures in multivariate biosignals, and (3) proposing an effective approach to combining S4, GSL, and GNN for modeling multivariate biosignals. Our main contributions are:

\begin{itemize}
    \item We propose \textsc{GraphS4mer} (Figure \ref{fig:model}), a general end-to-end GNN for modeling multivariate biosignals. Our modeling contributions are: (1) we leverage S4 to capture \textit{long-range temporal dependencies} in biosignals, (2) our model is able to \textit{dynamically learn the underlying graph structures} in the data without predefined graphs, and (3) our approach is a novel, effective way of combining S4, GSL, and GNN.
    \item We evaluate \textsc{GraphS4mer} on three datasets with distinct data modalities and tasks. Our model consistently outperforms existing methods on (1) seizure detection from EEG signals, outperforming a previous GNN with self-supervised pre-training by 3.1 points in AUROC; (2) sleep staging from polysomnographic signals, outperforming existing sleep staging models by 4.1 points in macro-F1 score; (3) ECG classification, outperforming previous state-of-the-art models by 2.7 points in macro-F1 score.
    \item Qualitative interpretability analysis suggests that our GSL method learns meaningful graph structures that reflect the underlying seizure classes, sleep stages, and ECG abnormalities.
\end{itemize}

\section{Related Works}
\textbf{Temporal graph neural networks.}
Temporal GNNs have been widely used for modeling multivariate time series. A recent work by \citet{Gao2022} shows that existing temporal GNNs can be grouped into two categories: time-and-graph and time-then-graph. The time-and-graph framework treats a temporal graph as a sequence of graph snapshots, constructs a graph representation for each snapshot, and embeds the temporal evolution using sequence models. For example, GCRN \citep{Seo2018} uses spectral graph convolution \citep{Defferrard2016} for node representations and Long Short-Term Memory network (LSTM) \citep{Hochreiter1997} for temporal dynamics modeling. DCRNN \citep{li2018-dcrnn} integrates diffusion convolution with Gated Recurrent Units (GRU) \citep{Cho2014-gru} for traffic forecasting. \citet{Li2019-path} combines relational graph convolution with LSTM for path failure prediction. EvolveGCN \citep{Pareja2020-evolvegcn} treats graph convolutional network (GCN) parameters as recurrent states and uses a recurrent neural network (RNN) to evolve the GCN parameters. In contrast, time-then-graph framework sequentially represents the temporal evolution of node and edge attributes, and uses these temporal representations to build a static graph representation \citep{Xu2020-Inductive, Rossi2020}. \citet{Gao2022} proposes a general framework for time-then-graph, which represents the evolution of node and edge features using two RNNs independently, constructs a static graph from these representations, and encodes them using a GNN. However, most of the existing temporal GNNs are limited to short sequences and adopt RNNs or self-attention for temporal modeling, which may be suboptimal when applied to long sequences. To our knowledge, only one study has leveraged S4 for temporal graphs \citep{El-Gazzar2022-graphs4}. However, the study is limited to fMRI data and the learned graph is static and shared across all fMRI data.

\textbf{Graph structure learning.}
GSL aims to jointly learn a graph structure and its corresponding node (or graph) representations. Briefly, GSL methods fall into the following three categories \citep{Zhu2021-gsl-survey}: (1) metric-based approaches that derive edge weights from node embeddings based on a metric function \citep{Li2018-agcn, Chen2020-iterative, Zhang2020-jt}, (2) neural-based approaches that use neural networks to learn the graph structure from node embeddings \citep{Luo2021-yy, Zheng2020-neuralsparse, jia2020-graphsleepnet}, and (3) direct approaches that treat the adjacency matrix or node embedding dictionary as learnable parameters \citep{Jin2020-prognn, Gao2020-rw}. GSL has been applied to temporal graph data. \citet{Zhang2020-cl} proposes learning graph adjacency matrix from traffic data through learnable parameters. \citet{Tang2022-traffic} uses a multi-layer perceptron and a similarity measure to learn the graph structure for traffic forecasting. \citet{shang2021-discrete} leverages a link predictor (two fully connected layers) to obtain the edge weights. Several works adaptively learn the adjacency matrix from data using learnable node embedding dictionaries \citep{Wu2019-graph-wavenet, El-Gazzar2021, El-Gazzar2022-graphs4}. However, these GSL approaches are limited to short sequences of less than 1k time steps and do not capture dynamically evolving graph structures.

\section{Background}
\subsection{Structured State Space model (S4)}
The Structured State Space sequence model (S4) \citep{Gu2022-s4} is a deep sequence model that is capable of capturing long-range dependencies in long sequences. S4 has outperformed prior state-of-the-art methods on pixel-level 1-D image classification, the challenging Long Range Arena benchmark for long sequence modeling, raw speech classification, and audio generation \citep{Gu2022-s4, goel22a-audio}.

S4 is based on state space model (SSM). A recent line of work \citep{Gu2020-hippo, Gu2021-lssl, Gu2022-s4} has shown that SSMs can be viewed as both CNNs and RNNs. The continuous-time SSM maps an 1-D signal $u(t)$ to a high dimensional latent state $x(t)$ before projecting to an 1-D output signal $y(t)$:
\begin{equation}
    x'(t) = \mathbf{A}x(t) + \mathbf{B}u(t), \ \
    y(t) = \mathbf{C}x(t) + \mathbf{D}u(t)
\label{eqn:cont_ssm}
\end{equation}

Equation \ref{eqn:cont_ssm} can be used as a black-box sequence representation in a deep learning model, where $\mathbf{A}$, $\mathbf{B}$, $\mathbf{C}$, and $\mathbf{D}$ are parameters learned by gradient descent. In the rest of this section, $\mathbf{D}$ is omitted for exposition because $\mathbf{D}u(t)$ can be viewed as a skip connection and can be easily computed \citep{Gu2022-s4}.

\citet{Gu2021-lssl} shows that discrete-time SSM is equivalent to the following convolution:
\begin{equation}
    \overline{\mathbf{K}} := (\overline{\mathbf{CB}}, \overline{\mathbf{CAB}}, ..., \overline{\mathbf{CA}}^{L-1} \overline{\mathbf{B}}), \ \ y = \overline{\mathbf{K}} * u
\end{equation}
where $\overline{\mathbf{A}}$, $\overline{\mathbf{B}}$, and $\overline{\mathbf{C}}$ are the discretized matrices of $\mathbf{A}, \mathbf{B}$, and $\mathbf{C}$, respectively; $\overline{\mathbf{K}}$ is called the SSM convolution kernel; and $L$ is the sequence length. 

S4 \citep{Gu2022-s4} is a special instantiation of SSMs that parameterizes $\mathbf{A}$ as a diagonal plus low-rank (DPLR) matrix. This parameterization has two major advantages. First, it enables fast computation of the SSM kernel $\overline{\mathbf{K}}$. Second, the parameterization includes the HiPPO matrices \citep{Gu2020-hippo}, which are a class of matrices that is capable of capturing long-range dependencies.

However, naively applying S4 for multivariate signals projects the signal channels to the hidden dimension with a linear layer \citep{Gu2022-s4}, which does not take into account the underlying graph structure of multivariate biosignals. This motivates our development of \textsc{GraphS4mer} described in Section \ref{methods}.

\subsection{Graph regularization}
\label{section:background_graph_reg}

Graph regularization encourages a learned graph to have several desirable properties, such as smoothness, sparsity, and connectivity \citep{Chen2020-iterative, Kalofolias2016-jt, Zhu2021-gsl-survey}. Let $\mathbf{X} \in \mathbb{R}^{N \times D}$ be a graph data with $N$ nodes and $D$ features. First, a common assumption in graph signal processing is that features change smoothly between adjacent nodes \citep{Ortega2018-gsp-review}. For instance, in an EEG graph constructed based on the standard EEG placement, neighboring nodes are physically adjacent EEG sensors, and their signals tend to have similar values because these adjacent EEG sensors measure brain electrical activity from adjacent brain areas. Given an undirected graph with adjacency matrix $\mathbf{W}$, the smoothness of the graph can be measured by the Dirichlet energy \citep{Belkin2001-lx}: 
\begin{align}
    \mathcal{L}_{\text{smooth}}(\mathbf{X}, \mathbf{W}) &= \frac{1}{2N^2} \sum_{i,j} \mathbf{W}_{ij} ||\mathbf{X}_{i,:} - \mathbf{X}_{j,:}||^2 \\ &= 
    \frac{1}{N^2} \text{tr}\big(\mathbf{X}^T \mathbf{L} \mathbf{X} \big)
\label{eqn:smoothness}
\end{align}
where tr($\cdot$) denotes the trace of a matrix, $\mathbf{L} = \mathbf{D} - \mathbf{W}$ is the graph Laplacian, $\mathbf{D}$ is the degree matrix of $\mathbf{W}$. In practice, the normalized graph Laplacian $\hat{\mathbf{L}} = \mathbf{D}^{-1/2}\mathbf{L}\mathbf{D}^{-1/2}$ is used so that the Dirichlet energy is independent of node degrees. Intuitively, the eigenvectors of the normalized graph Laplacian are minimizers of the Dirichlet energy \citep{cai2020note}, and thus minimizing Equation \ref{eqn:smoothness} encourages the learned graph to be smooth.

However, simply minimizing the Dirichlet energy may result in a trivial solution $\mathbf{W} = \mathbf{0}$, i.e., $\mathcal{L}_{\text{smooth}}=0$ if $\mathbf{W}=\mathbf{0}$. To avoid this trivial solution and encourage sparsity of the graph, additional constraints can be added \citep{Chen2020-iterative, Kalofolias2016-jt}:
\begin{align}
    \mathcal{L}_{\text{degree}}(\mathbf{W}) &= -\frac{1}{N} \mathbf{1}^T \log(\mathbf{W}\mathbf{1})
    \label{eqn:degree} \\
    \mathcal{L}_{\text{sparse}}(\mathbf{W}) &= \frac{1}{N^2} ||\mathbf{W}||_{F}^2 
    \label{eqn:sparse}
\end{align}

where $||.||_F$ is the Frobenius norm of a matrix. Intuitively, $\mathcal{L}_{\text{degree}}$ penalizes disconnected graphs and $\mathcal{L}_{\text{sparse}}$ discourages nodes with high degrees (i.e., encourages the learned graph to be sparse).

\section{Methods}
\label{methods}
Figure \ref{fig:model} illustrates the architecture of \textsc{GraphS4mer}. It has three main components: (1) stacked S4 layers, (2) a GSL layer, and (3) GNN layers. Sections \ref{section:methods_problem_setup}--\ref{section:methods_model} describe the details of the problem setup and model architecture.

\subsection{Problem setup}
\label{section:methods_problem_setup}
To capture non-Euclidean correlations among sensors, we propose a general representation of multivariate biosignals using graphs. Let $\mathbf{X} \in \mathbb{R}^{N \times T \times M}$ be a multivariate biosignal, where $N$ is the number of sensors, $T$ is the sequence length, and $M$ is the input dimension of the signal (typically $M = 1$). We represent the biosignal as a graph $\mathcal{G} = \{\mathcal{V}, \mathcal{E}, \mathbf{W} \}$, where the set of nodes $\mathcal{V}$ corresponds to the sensors (channels/leads), $\mathcal{E}$ is the set of edges, and $\mathbf{W}$ is the adjacency matrix. Here, $\mathcal{E}$ and $\mathbf{W}$ are unknown and will be learned by our model.

While our formulation is general to any type of node or graph classification and regression tasks, we focus on graph classification tasks in this work. Graph classification aims to learn a function which maps an input signal to a prediction, i.e., $f: \mathbf{X} \rightarrow y \in \{1, 2, ..., C\}$, where $C$ is the number of classes.

\subsection{Temporal modeling with S4}
\label{section:methods_s4}
As biosignals often involve long-range temporal correlations, we leverage S4 to capture these long-range temporal dependencies. However, naively applying S4 for multivariate signals projects the $N$ signal channels to the hidden dimension with a linear layer \citep{Gu2022-s4}, which may be suboptimal because it neglects the underlying graph structure of biosignals. Instead, we use stacked S4 layers with residual connection to embed signals in each channel (sensor) independently, resulting in an embedding $\mathbf{H} \in \mathbb{R}^{N \times T \times D}$ (referred to as ``S4 embeddings" hereafter) for each input signal $\mathbf{X}$, where $D$ is the embedding dimension.

\subsection{Dynamic graph structure learning}
\label{section:gsl}
\textbf{Graph structure learning layer.} As motivated in Introduction, the graph structure of multivariate biosignals may not be easily defined due to unknown sensor locations, and the underlying graph connectivity can evolve over time. Hence, we develop a graph structure learning (GSL) layer to learn a \textit{unique} graph structure within a short time interval $r$, where $r$ is a pre-specified resolution. Instead of learning a unique graph at each time step, we choose to learn a graph over a time interval of length $r$ because (1) aggregating information across a longer time interval can result in less noisy graphs and (2) it is more computationally efficient. For convenience, we let the predefined resolution $r$ be an integer and assume that the sequence length $T$ is divisible by $r$ without overlaps. In the remainder of this paper, we denote $n_d = \frac{T}{r}$ as the number of dynamic graphs.


For graph classification problems, the model's task is to predict the classes of unseen multivariate biosignals. Therefore, the goal of GSL is to learn a unique graph for each biosignal within $r$ time intervals. Moreover, the GSL layer is expected to be able to generalize to unseen signals. Learning the graph structure is equivalent to learning to compute the edge weight between each pair of nodes. We observe that there is an analogy between self-attention \citep{Vaswani2017-transformer} in natural language processing and GSL---self-attention results in an attention weight between each pair of words in a sentence, and GSL learns the edge weight between each pair of nodes. Given this analogy, we adopt self-attention \citep{Vaswani2017-transformer} to allow each node to attend to all other nodes, and we use the attention weights as the edge weights. The adjacency matrix of the $t$-th dynamic graph ($t=1,2,...,n_d$), $\overline{\mathbf{W}}^{(t)} \in \mathbb{R}^{N \times N}$, is learned as follows:
\begin{align}
    \label{eqn:gsl_inductive}
    \mathbf{Q} &= \mathbf{h}^{(t)} \mathbf{M}^Q, \ \ \mathbf{K} = \mathbf{h}^{(t)} \mathbf{M}^K, \\
    \overline{\mathbf{W}}^{(t)} &= \text{softmax}(\frac{\mathbf{Q} \mathbf{K}^T}{\sqrt{D}}), \ \text{for} \ t = 1, 2, ..., n_d
\end{align}
Here, $\mathbf{h}^{(t)} \in \mathbb{R}^{N \times D}$ is mean-pooled S4 embeddings within the $t$-th time interval of length $r$; $\mathbf{M}^Q \in \mathbb{R}^{D \times D}$ and $\mathbf{M}^K \in \mathbb{R}^{D \times D}$ are weights projecting $\mathbf{h}^{(t)}$ to query $\mathbf{Q}$ and key $\mathbf{K}$, respectively. The above equations can be easily extended to multihead self-attention \citep{Vaswani2017-transformer}.

There are two major differences between our GSL layer and prior works \citep{Wu2019-graph-wavenet, El-Gazzar2021, El-Gazzar2022-graphs4}. First, our GSL layer is applied to S4 embeddings rather than raw signals, which leverages the long-range dependencies learned by S4 layers. Second, our GSL layer is able to learn dynamically evolving graph structures over time.

To guide the graph structure learning process, for the $t$-th dynamic graph, we add a k-nearest neighbor (KNN) graph $\mathbf{W}_{\text{KNN}}^{(t)}$ to the learned adjacency matrix $\overline{\mathbf{W}}^{(t)}$, where each node's k-nearest neighbors are defined by cosine similarity between their respective values in $\mathbf{h}^{(t)}$ \citep{Chen2020-iterative}:
\begin{align}
  \mathbf{W}^{(t)} = \epsilon \mathbf{W}_{\text{KNN}}^{(t)} + (1 - \epsilon) \overline{\mathbf{W}}^{(t)}
\label{eqn:add_knn_graph}
\end{align}
Here, $\epsilon \in [0, 1)$ is a hyperparameter for the weight of the KNN graph.

To introduce graph sparsity for computational efficiency, we prune the adjacency matrix $\mathbf{W}^{(t)}$ by removing edges whose weights are smaller than a certain threshold \citep{Chen2020-iterative}, i.e., $\mathbf{W}_{ij}^{(t)} = 0$ if $\mathbf{W}_{ij}^{(t)} < \kappa$, where $\kappa$ is a hyperparameter. In this study, we treat the biosignals as undirected, as each pair of sensors is correlated (or anti-correlated) without directionality. To ensure that the learned graphs are undirected, we make the learned adjacency matrix symmetric by taking the average of the edge weights between two nodes. 

\textbf{Graph regularization.} To encourage the learned graphs to have desirable graph properties, we include three regularization terms commonly used in the GSL literature \citep{Chen2020-iterative, Kalofolias2016-jt, Zhu2021-gsl-survey} (see Section \ref{section:background_graph_reg} and Eqns \ref{eqn:smoothness}--\ref{eqn:sparse}). The regularization loss is the weighted sum of the three regularization terms and averaged across all dynamic graphs: $\mathcal{L}_{\text{reg}} = \frac{1}{n_d} \sum_{t=1}^{n_d} \alpha \mathcal{L}_{\text{smooth}}(\mathbf{h}^{(t)}, \mathbf{W}^{(t)}) + \beta \mathcal{L}_{\text{degree}}(\mathbf{W}^{(t)}) +  \gamma \mathcal{L}_{\text{sparse}}(\mathbf{W}^{(t)})$, where $\alpha$, $\beta$, and $\gamma$ are hyperparameters.

\subsection{Model architecture}
\label{section:methods_model}
The overall architecture of our model (Figure \ref{fig:model} and Algorithm \ref{pseudocode} in Appendix \ref{supp:pseudocode}), \textsc{GraphS4mer}, consists of three main components: (1) stacked S4 layers with residual connection to model temporal dependencies in signals within each sensor independently, which maps raw signals $\mathbf{X} \in \mathbb{R}^{N \times T \times M}$ to S4 embedding $\mathbf{H} \in \mathbb{R}^{N \times T \times D}$, (2) a GSL layer to learn dynamically evolving adjacency matrices $\mathbf{W}^{(1)}, ..., \mathbf{W}^{(n_d)}$, and (3) GNN layers to learn spatial dependencies between sensors given the learned graph structures $\mathbf{W}^{(1)}, ..., \mathbf{W}^{(n_d)}$ and node features $\mathbf{H}$. While our model is general to any kind of GNN layers, we use an expressive GNN architecture---Graph Isomorphism Network \citep{Xu2018-gin, Hu2020-strategies}---in our experiments.

Finally, a temporal pooling layer and a graph pooling layer are added to aggregate temporal and spatial representations, respectively, which is followed by a fully connected layer to produce a prediction for each multivariate biosignal. The total loss is the sum of the graph regularization loss and the classification loss. Figure \ref{fig:model} shows the overall model architecture, and Algorithm \ref{pseudocode} in Appendix \ref{supp:pseudocode} shows the pseudocode for \textsc{GraphS4mer}.

\section{Experiments}

\subsection{Experimental setup}
In this section, we briefly introduce the datasets and experimental setup in our study. For each experiment, we ran three runs with different random seeds and report mean and standard deviation of the results. See Appendix \ref{supp:dataset_details}--\ref{supp:model_training} for details on datasets, baseline models, model training procedures, and hyperparameters.

\textbf{Seizure detection from EEGs.} We first evaluate our model on EEG-based seizure detection. We use the publicly available Temple University Hospital Seizure Detection Corpus (TUSZ) v1.5.2 \citep{Shah2018}, and follow the same experimental setup of seizure detection on 60-s EEG clips as in \citet{Tang2022-gnn-ssl}. The number of EEG sensors is 19. The sampling rate of each 60-s EEG clip is 200 Hz, resulting in a sequence length of 12k time steps. The task is binary classification of detecting whether or not a 60-s EEG clip contains seizure. The resolution $r$ in GSL layer is chosen as 10-s (i.e., 2,000 time steps), which is inspired by how trained EEG readers analyze EEGs. Following prior studies, we use AUROC, F1-score, area under precision-recall curve (AUPRC), sensitivity, and specificity as the evaluation metrics.

\textbf{Sleep staging from polysomnographic signals.} Next, we evaluate our model on sleep staging from polysomnographic (PSG) signals. PSG is used in the diagnosis of sleep disorders such as obstructive sleep apnea. We use the publicly available Dreem Open Dataset-Healthy (DOD-H) \citep{Guillot2020-dreem}. The number of PSG sensors is 16. The sampling rate is 250 Hz, and thus each 30-s signal has 7,500 time steps. The task is to classify each 30-s PSG signal as one of the five sleep stages: wake, rapid eye movement (REM), non-REM sleep stages N1, N2, and N3. The resolution $r$ is 10-s (i.e., 2.5k time steps). Similar to a prior study \citet{Guillot2021-robustsleepnet}, we use macro-F1 score and Cohen's Kappa as the evaluation metrics.

\textbf{ECG classification.} Lastly, we evaluate our model on a 12-lead ECG classification task. We use the publicly available ICBEB ECG dataset \citep{Liu2018-pu}, and follow the same data split as described in \citet{Strodthoff2021-tx}. The number of ECG channels is 12, with a sampling rate of 100 Hz. The ECG lengths vary from 6-s to 60-s, resulting in variable sequence lengths between 600 and 6k time steps. To allow training in mini-batches across variable sequence lengths, we pad short sequences with 0s and mask out the padded values so that the padded values are not seen by the models. There are 9 classes in the dataset (see Table \ref{supp_tab:icbeb_data} in Appendix \ref{supp:dataset_details}), including one normal class and 8 abnormal classes indicating diseases. Each ECG is labeled with at least one of the 9 classes; therefore, the task is a multilabel classification task. The resolution $r$ in the GSL layer is set as the actual sequence length of each ECG to allow training in batches. Following prior studies \citep{Liu2018-pu, Strodthoff2021-tx}, we use macro-F1, macro-F2, macro-G2, and macro-AUROC as the evaluation metrics. See Appendix \ref{supp:icbeb_eval_metrics} for detailed definitions of these metrics.

\subsection{Experimental results}

\textbf{Seizure detection from EEG signals.} Table \ref{tab:tuh_results} shows our model performance on EEG-based seizure detection and its comparison to existing models. \textsc{GraphS4mer} outperforms the previous state-of-the-art, Dist-DCRNN with pre-training, by 3.1 points in AUROC. Notably, Dist-DCRNN was pre-trained using a self-supervised task \citep{Tang2022-gnn-ssl}, whereas \textsc{GraphS4mer} was trained from scratch without the need of pre-training. 

\begin{table*}[ht]
\centering
\caption{\textbf{Results of EEG-based seizure detection on TUSZ dataset.} Baseline model results are cited from \citet{Tang2022-gnn-ssl}. Best and second best results are \textbf{bolded} and \underline{underlined}, respectively.}
\resizebox{\textwidth}{!}{\begin{tabular}{lccccc}
\toprule
Model                       & AUROC         & F1-Score      & AUPRC         & Sensitivity   & Specificity   \\ \midrule
LSTM \citep{Hochreiter1997}                        & 0.715 $\pm$ 0.016 & 0.365 $\pm$ 0.009 & 0.287 $\pm$ 0.026 & 0.463 $\pm$ 0.060 & 0.814 $\pm$ 0.053 \\
Dense-CNN \citep{Saab2020}                   & 0.796 $\pm$ 0.014 & 0.404 $\pm$ 0.022 & 0.399 $\pm$ 0.017 & 0.451 $\pm$ 0.134 & 0.869 $\pm$ 0.071 \\
CNN-LSTM \citep{Ahmedt2020-cnnlstm}                    & 0.682 $\pm$ 0.003 & 0.330 $\pm$ 0.016 & 0.276 $\pm$ 0.009 & 0.363 $\pm$ 0.044 & 0.857 $\pm$ 0.023 \\
Dist-DCRNN w/o Pre-training \citep{Tang2022-gnn-ssl} & 0.793 $\pm$ 0.022 & 0.341 $\pm$ 0.170 & 0.418 $\pm$ 0.046 & 0.326 $\pm$ 0.183 & \underline{0.932 $\pm$ 0.058} \\
Corr-DCRNN w/o Pre-training \citep{Tang2022-gnn-ssl} & 0.804 $\pm$ 0.015 & 0.448 $\pm$ 0.029 & 0.440 $\pm$ 0.021 & 0.457 $\pm$ 0.058 & 0.900 $\pm$ 0.028 \\
Dist-DCRNN w/ Pre-training \citep{Tang2022-gnn-ssl} & \underline{0.875 $\pm$ 0.016} & \underline{0.571 $\pm$ 0.029} & \underline{0.593 $\pm$ 0.031} & \underline{0.570 $\pm$ 0.047} & 0.927 $\pm$ 0.012 \\
Corr-DCRNN w/ Pre-training \citep{Tang2022-gnn-ssl}  & 0.850 $\pm$ 0.014 & 0.514 $\pm$ 0.028 & 0.539 $\pm$ 0.024 & 0.502 $\pm$ 0.047 & 0.923 $\pm$ 0.008 \\ \midrule
\textsc{GraphS4mer} (ours)           & \textbf{0.906 $\pm$ 0.012} & \textbf{0.680 $\pm$ 0.012}  & \textbf{0.723 $\pm$ 0.023}  & \textbf{0.718 $\pm$ 0.041}  & \textbf{0.933 $\pm$ 0.009} \\
\bottomrule
\end{tabular}}
\label{tab:tuh_results}
\end{table*}

\textbf{Sleep staging from PSG signals.} Table \ref{tab:dodh_results} compares our model performance to existing sleep staging models on DOD-H dataset. \textsc{GraphS4mer} outperforms RobustSleepNet, a specialized sleep staging model, by 4.1 points in macro-F1. Note that the baselines preprocess the PSG signals using short-time Fourier transform, whereas our model directly takes in raw PSG signals without the need of preprocessing.

\begin{table*}[h!]
\centering
\caption{\textbf{Results of sleep staging on DOD-H dataset.} Best and second best results are \textbf{bolded} and \underline{underlined}, respectively.}
\resizebox{0.6\textwidth}{!}{\begin{tabular}{lcc}
\toprule
Model             & Macro-F1     & Kappa        \\
\midrule
LSTM \citep{Hochreiter1997}              &   0.609 $\pm$ 0.034  & 0.539 $\pm$ 0.046  \\
SimpleSleepNet \citep{Guillot2020-dreem}   & 0.720 $\pm$ 0.001 & 0.703 $\pm$ 0.013 \\
RobustSleepNet \citep{Guillot2021-robustsleepnet}    & \underline{0.777 $\pm$ 0.007} & \underline{0.758 $\pm$ 0.008} \\
DeepSleepNet \citep{Supratak2017-td}     & 0.716 $\pm$ 0.025 & 0.711 $\pm$ 0.032 \\ \midrule
\textsc{GraphS4mer} (ours) & \textbf{0.818 $\pm$ 0.008} & \textbf{0.802 $\pm$ 0.014} \\
\bottomrule
\end{tabular}}
\label{tab:dodh_results}
\end{table*}

\textbf{ECG classification.} Table \ref{tab:icbeb_results} compares \textsc{GraphS4mer} to existing ECG classification models on ICBEB dataset. \textsc{GraphS4mer} provides 2.7 points improvement in macro-F1 score compared to XResNet1D \citep{He2019-rt}, a prior state-of-the-art CNN specialized in time series modeling. Note that all the baselines are trained on 2.5-s ECG windows (i.e., 250 time steps) and the window-wise predictions are aggregated post-hoc to obtain the predictions for each ECG, whereas our model directly takes the entire ECG signal (ranging from 600 to 6k time steps) without the need of sliding windows, which allows modeling long-range temporal dependencies in the entire ECG. Table \ref{supp_tab:ecg_indiv_class} in Appendix \ref{supp:ecg_indiv_classes} shows model performance in individual ECG classes. \textsc{GraphS4mer} outperforms the baselines on 6 out of 9 classes.

\begin{table*}[h!]
\centering
\caption{\textbf{Results of ECG classification on ICBEB dataset.} Best and second best results are \textbf{bolded} and \underline{underlined}, respectively.}
\resizebox{0.87\textwidth}{!}{\begin{tabular}{lcccc}
\toprule
Model         & Macro F1-Score & Macro F2-Score & Macro G2-Score & Macro AUROC  \\ \midrule
InceptionTime \citep{Ismail_Fawaz2020-dh} & 0.778 $\pm$ 0.020   & 0.792 $\pm$ 0.023   & 0.576 $\pm$ 0.027   & 0.964 $\pm$ 0.008 \\
XResNet1D \citep{He2019-rt}     & \underline{0.782 $\pm$ 0.016}  & \underline{0.803 $\pm$ 0.016}   & \underline{0.587 $\pm$ 0.019}   & \underline{0.971 $\pm$ 0.001} \\
ResNet1D \citep{Wang2017-sc}      & 0.772 $\pm$ 0.015   & 0.788 $\pm$ 0.008   & 0.570 $\pm$ 0.014   & 0.963 $\pm$ 0.004 \\
FCN \citep{Wang2017-sc}         & 0.736 $\pm$ 0.015   & 0.764 $\pm$ 0.014   & 0.538 $\pm$ 0.014   & 0.950 $\pm$ 0.003 \\
LSTM \citep{Hochreiter1997}          & 0.747 $\pm$ 0.018   & 0.768 $\pm$ 0.018   & 0.542 $\pm$ 0.018   & 0.942 $\pm$ 0.004 \\
Bidir-LSTM \citep{Hochreiter1997}    & 0.748 $\pm$ 0.009   & 0.769 $\pm$ 0.004   & 0.548 $\pm$ 0.005   & 0.945 $\pm$ 0.002 \\
WaveletNN \citep{Strodthoff2021-tx}    & 0.621 $\pm$ 0.013   & 0.643 $\pm$ 0.019   & 0.396 $\pm$ 0.014   & 0.911 $\pm$ 0.002 \\ \midrule
\textsc{GraphS4mer} (ours)    & \textbf{0.809 $\pm$ 0.004}   & \textbf{0.804 $\pm$ 0.004}   & \textbf{0.609 $\pm$ 0.005}   & \textbf{0.977 $\pm$ 0.001} \\ \bottomrule
\end{tabular}}
\label{tab:icbeb_results}
\end{table*}

\textbf{Ablations.} We perform ablation studies to investigate the importance of (1) graph-based representation, where we remove the GSL and GNN layers in our model; (2) S4 encoder, where we replace the S4 layers with GRUs \citep{Cho2014-gru}; (3) long-range temporal modeling for GSL and graph representation learning, where we apply GSL and GNN layers to raw signals, followed by S4 layers; and (4) GSL, where we remove the GSL layer and use predefined graphs. For seizure detection on TUSZ dataset, we use the predefined distance-based EEG graph as in \citet{Tang2022-gnn-ssl}. DOD-H and ICBEB datasets have no predefined graph available, and thus we use a KNN graph obtained based on cosine similarity between raw signals in a pair of nodes (setting K the same as \textsc{GraphS4mer}). Table \ref{tab:ablation} shows the ablation results. There are several important observations. First, \textsc{GraphS4mer} consistently outperforms S4, suggesting the effectiveness of representing multivariate biosignals as graphs. Second, when S4 layers are replaced with GRUs, the model performance drops by a large margin. This indicates the effectiveness of S4 in modeling long-range temporal dependencies. Third, applying GSL and GNN layers before S4 layers (``GSL-GNN-S4") results in lower performance than \textsc{GraphS4mer} on all datasets, suggesting the effectiveness of capturing long-range temporal dependencies for GSL and graph representation learning. Finally, \textsc{GraphS4mer} with GSL outperforms \textsc{GraphS4mer} without GSL (with predefined graph) for sleep staging on the DOD-H dataset, and performs marginally better than \textsc{GraphS4mer} without GSL for seizure detection on the TUSZ dataset and ECG classification on the ICBEB dataset. Nevertheless, a major advantage of our GSL method is that it does not require prior knowledge about the graph structure, which would be particularly useful when sensor locations are not available. Table \ref{supp_tab:ablation_more_metrics} in Appendix \ref{supp:ablation_more_metrics} shows additional evaluation metrics for the ablation experiments.

\begin{table*}[h!]
\centering
\caption{\textbf{Ablation results.} For S4, we remove the GSL and GNN layers. For \textsc{GraphS4mer} w/o S4, we replace S4 with GRUs. For GSL-GNN-S4, we apply GSL and GNN layers to raw signals, followed by S4 layers. For \textsc{GraphS4mer} w/o GSL, we remove the GSL layer and use predefined graphs. P-values are computed using bootstrapping with 5,000 replicates with replacement on the run with median performance. Best and second best results are \textbf{bolded} and \underline{underlined}, respectively.}
\resizebox{0.77\textwidth}{!}{\begin{tabular}{l|cc|cc|cc}
\toprule
\multirow{2}{*}{Model} & \multicolumn{2}{c|}{TUSZ} & \multicolumn{2}{c|}{DOD-H} & \multicolumn{2}{c}{ICBEB} \\ 
                       & AUROC          & p-value & Macro-F1       & p-value  & Macro-F1       & p-value  \\ \midrule
S4                     & 0.824 $\pm$ 0.011  & 0.0002  & 0.778 $\pm$ 0.009   & 0.0004   & 0.781 $\pm$ 0.003   & 0.0132   \\
\textsc{GraphS4mer} w/o S4      & 0.705 $\pm$ 0.095   & 0.0002  & 0.634 $\pm$ 0.061   & 0.0002   & 0.197 $\pm$ 0.005   & 0.0002   \\
GSL-GNN-S4             & 0.882 $\pm$ 0.014   & 0.0004  & \underline{0.797 $\pm$ 0.011}   & 0.0164   & 0.772 $\pm$ 0.012   & 0.0060    \\
\textsc{GraphS4mer} w/o GSL     & \underline{0.899 $\pm$ 0.010}  & 0.7864  & 0.765 $\pm$ 0.016   & 0.0002   & \underline{0.797 $\pm$ 0.012}   & 0.1268   \\ \midrule
\textsc{GraphS4mer} (ours)      & \textbf{0.906 $\pm$ 0.012}  & ref     & \textbf{0.818 $\pm$ 0.008}   & ref      & \textbf{0.809 $\pm$ 0.004}   & ref    \\ \bottomrule
\end{tabular}}
\label{tab:ablation}
\end{table*}

\textbf{Effect of temporal resolution.} We examine the effect of temporal resolution $r$ in Figure \ref{supp_fig:effect_of_r} in Appendix \ref{supp:effect_of_r} for seizure detection and sleep staging tasks. For ECG classification, the temporal resolution $r$ is set as the actual sequence lengths due to variable lengths of the ECG signals, and thus is excluded for this analysis. We observe that a smaller value of $r$ tends to result in higher performance than a larger value of $r$, suggesting the effectiveness of learning dynamically varying graphs.

\begin{figure*}[h!]
    \centering
    \includegraphics[width=\textwidth]{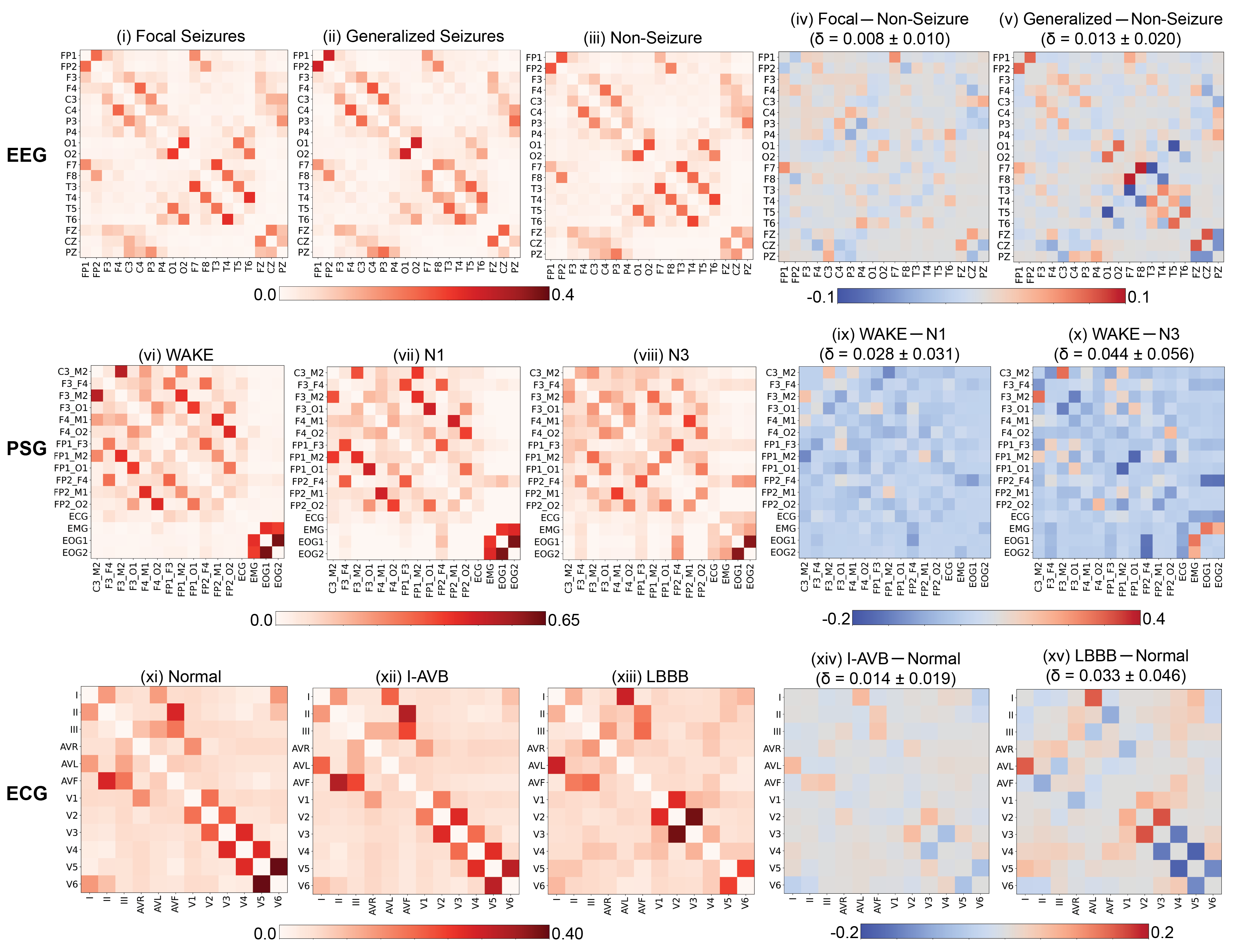}
    \caption{\textbf{Top}: Mean adjacency matrix for EEG in correctly predicted test samples for \textbf{(i)} focal seizures, \textbf{(ii)} generalized seizures, and \textbf{(iii)} non-seizure. \textbf{(iv)} Difference between focal seizure and non-seizure adjacency matrices. \textbf{(v)} Difference between generalized seizure and non-seizure adjacency matrices. \textbf{Middle}: Mean adjacency matrix for PSG in correctly predicted test samples for sleep stage \textbf{(vi)} WAKE, \textbf{(vii)} N1, and \textbf{(viii)} N3. \textbf{(ix)} Difference between WAKE and N1 adjacency matrices. \textbf{(x)} Difference between WAKE and N3 adjacency matrices. \textbf{Bottom}: Mean adjacency matrix for ECG in correctly predicted test samples for \textbf{(xi)} normal ECG, \textbf{(xii)} first-degree atrioventricular block (I-AVB), and \textbf{(xiii)} left bundle branch block (LBBB). \textbf{(xiv)} Difference between I-AVB and normal adjacency matrices. \textbf{(xv)} Difference between LBBB and normal adjacency matrices. $\delta$: mean and std of absolute values of differences between two mean adjacency matrices.}
\label{fig:viz_eeg_sleep_adj}
\end{figure*}

\textbf{Effect of KNN graph weight and graph regularization.} In Figures \ref{supp_fig:effect_of_knn}--\ref{supp_fig:effect_of_graph_reg} in Appendix \ref{supp:effect_of_knn_graph_reg}, we examine the effect of KNN graph weight (i.e., $\epsilon$ in Eqn \ref{eqn:add_knn_graph}) and graph regularization on model performance. Larger KNN graph weights and graph regularization result in marginally better performance.

\textbf{Computational costs of GSL.} In Appendix \ref{supp:comp_cost} and Table \ref{supp_tab:comp_cost}, we examine the computational costs of \textsc{GraphS4mer} across different values of the time interval $r$ in the GSL layer (see Section \ref{section:gsl}). As the time interval $r$ decreases (i.e., number of dynamic graphs increases), the number of model parameters remains the same, whereas the number of multiply-accumulate (MACs) operations increases. This is expected given that the same GSL layer is applied multiple times to construct dynamic graphs.

\textbf{Interpretation of graphs.} To investigate if the learned graphs are meaningful, we visualize mean adjacency matrices for EEG, PSG, and ECG signals in the correctly predicted test samples in Figure \ref{fig:viz_eeg_sleep_adj}, grouped by seizure classes, sleep stages, and ECG classes, respectively. To quantify the differences between two mean adjacency matrices, we report the mean and standard deviation of the absolute values of the differences in Figure \ref{fig:viz_eeg_sleep_adj} (referred to as $\delta$). These adjacency matrices were reviewed and interpreted by clinical experts in terms of whether the differences in adjacency matrices between diseased (or sleep) and normal (or wake) classes reflect characteristics of the diseased (or sleep) states, and several clinically-meaningful patterns were found. 

For EEG, the magnitude of differences between generalized seizure and non-seizure adjacency matrices (Figure \ref{fig:viz_eeg_sleep_adj}v; $\delta = 0.013 \pm 0.020$) is larger than the magnitude of differences between focal seizure and non-seizure adjacency matrices (Figure \ref{fig:viz_eeg_sleep_adj}iv; $\delta = 0.008 \pm 0.010$). This suggests that the abnormalities in generalized seizures are more synchronized among the channels, resulting in high correlations. This is consistent with the literature that generalized seizures are characterized with abnormally synchronized brain activity \citep{Gloor1990-vv, Amor2009-tk}.

For PSG, we observe that the magnitude of differences between N3 and wake (Figure \ref{fig:viz_eeg_sleep_adj}x; $\delta=0.044 \pm 0.056$) is larger than the magnitude of differences between N1 and wake (Figure \ref{fig:viz_eeg_sleep_adj}ix; $\delta=0.028 \pm 0.031$). This pattern is expected given that N1 is the earliest sleep stage, whereas N3 is the deep sleep stage and is associated with slow brain waves that are not present in other sleep stages \citep{AASM2007}. Mean adjacency matrices for sleep stages N2 and REM are shown in Figure \ref{supp_fig:sleep_adj_all} in Appendix \ref{supp:sleep_adj_all}.

For ECG, we find that the magnitude of differences between adjacency matrices of first-degree atrioventricular block (I-AVB) and normal ECG is small (Figure \ref{fig:viz_eeg_sleep_adj}xiv; $\delta=0.014 \pm 0.019$), which is expected as I-AVB abnormality involves only small changes in the ECG (a subtle increase in the PR interval of ECG, not involving any morphologic changes in the p-waves or the QRS complexes). In contrast, the magnitude of differences between adjacency matrices of left bundle branch block (LBBB) and normal ECG is larger, particularly in ECG leads V1-V6 (Figure \ref{fig:viz_eeg_sleep_adj}xv; $\delta=0.033 \pm 0.046$). This finding is clinically meaningful as ECG signals with LBBB demonstrate a pronounced abnormality in the QRS complexes of the ECG, especially in leads V1-V6. In Appendix \ref{supp:ecg_adj_all}, we show example normal ECG, an ECG with I-AVB, and an ECG with LBBB in Figure \ref{supp_fig:ecg_examples}, as well as adjacency matrices for nine ECG classes in Figure \ref{supp_fig:ecg_adj_all}.

\section{Discussion and Limitations}
In this study, we propose a novel GSL method to learn dynamically evolving graph structures in multivariate biosignals, and present \textsc{GraphS4mer}, a general GNN integrating S4 and GSL for modeling multivariate biosignals. Our method sets new state-of-the-art performance on seizure detection, sleep staging, and ECG classification, and learns meaningful graph structures that reflect seizure classes, sleep stages, and ECG abnormalities.

In general, \textsc{GraphS4mer} is best suited for biosignals that (1) are measured by multiple sensors placed in or on the surface of a person's body and (2) exhibit long-range temporal dependencies. The three data types---EEG, PSG, and ECG---satisfy these properties \citep{Berthouze2010-qf, Peng1995-mg, Garcia2013-du, Sazgar2019-bd}, and \textsc{GraphS4mer} outperforms prior state-of-the-art methods on the related classification tasks.

Ablation experiments (Table \ref{tab:ablation}) suggest the importance of each component of \textsc{GraphS4mer}. First, \textsc{GraphS4mer} consistently outperforms S4, indicating the effectiveness of representing biosignals as graphs. Second, replacing S4 with GRU decreases the model performance by a large margin, which suggests the importance of S4 in modeling long-range temporal dependencies. Notably, the three tasks in our study cover a wide range of sequence lengths ranging from 600 to 12k time steps. Third, applying GSL and GNN layers prior to S4 layers underperforms \textsc{GraphS4mer}, which indicates that long-range temporal dependencies learned by S4 are crucial for learning graph structures and graph representations. Lastly, \textsc{GraphS4mer} outperforms \textsc{GraphS4mer} without GSL (with predefined graphs) on sleep staging, which suggests that GSL could be particularly useful for applications where graph structures cannot be easily defined.

There exists several limitations in this study. First, the GSL layer does not take into account domain knowledge. For example, GSL does not consider the standard EEG placement, and it only performs marginally better than the predefined EEG graph (Table \ref{tab:ablation}). Future work can leverage domain knowledge, such as standard sensor placement, to further improve the model performance. Second, self-attention requires time and memory that grows quadratically with the number of nodes, and thus could be suboptimal when the number of sensors is large. Future work can leverage more efficient attention mechanisms \citep{child2019generating, Kitaev2020Reformer, dao2022flashattention} to improve the computational efficiency of GSL. Third, due to variable ECG lengths in the ICBEB dataset, we set the temporal resolution in the GSL layer to the actual sequence length for each ECG to facilitate model training in batches. Future work can improve the model implementation to allow using shorter temporal resolutions for biosignals with variable lengths. Lastly, we only investigate biosignal classification tasks in this study. Future study can apply our model to other use cases.

\section{Conclusion}
In conclusion, we developed \textsc{GraphS4mer}, a general GNN for modeling multivariate biosignals. Our method outperformed prior work on seizure detection, sleep staging, and ECG classification, and learned meaningful graph structures that reflect the underlying classes.

\acks{We thank Stanford HAI and Google Cloud for the support of Google Cloud credits for this work.}

\bibliography{jmlr-sample}

\newpage
\newpage
\appendix

\section{Pseudocode for GraphS4mer}
\label{supp:pseudocode}

Algorithm \ref{pseudocode} shows the pseudocode for \textsc{GraphS4mer}. Note that $n_d = \frac{T}{r}$ is the number of dynamic graphs, where $T$ is the sequence length and $r$ is the pre-specified resolution. While there is a for loop over $t = 1, ..., n_d$, the code implementation can be done in batches to eliminate the need of a for loop.

\begin{algorithm2e}
\caption{Pseudocode for \textsc{GraphS4mer}.}
\label{pseudocode}
    \SetKwInOut{KwIn}{Input}
    \SetKwInOut{KwOut}{Output}
    \SetKwInOut{Parameter}{Parameter}
    \KwIn{multivariate biosignal $\mathbf{X}$, label $\mathbf{y}$}
    \Parameter{model weights $\mathbf{\theta}$, including weights for S4, GSL, GNN, and linear layers}
    \KwOut{prediction $\hat{\mathbf{y}}$, learned graph adjacency matrices $\mathbf{W}^{(1)}, ..., \mathbf{W}^{(n_d)}$}
    Randomly initialize model weights $\mathbf{\theta}$
    
    \While{not converged}{
    $\mathcal{L}_{\text{reg}} \leftarrow 0$
    
    \tcp{step 1. get S4 embeddings}
    
    $\mathbf{H}$ $\leftarrow$ S4($\mathbf{X}$)
    
    \For{$t = 1, 2, ..., n_d$}{
        
        $\mathbf{h}^{(t)} \leftarrow \text{mean-pool}(\mathbf{H}_{:,((t-1) \times r):(t \times r),:})$ \tcp{mean-pool along temporal dim}
        
        \tcp{step 2. graph structure learning (GSL)}
        Construct KNN graph, $\mathbf{W}_{\text{KNN}}^{(t)}$, based on cosine similarity between nodes using $\mathbf{h}^{(t)}$
        
        $\mathbf{W}^{(t)} \leftarrow \text{GSL}(\mathbf{h}^{(t)}, \mathbf{W}_{\text{KNN}}^{(t)})$ using Eqn. \ref{eqn:gsl_inductive} --\ref{eqn:add_knn_graph}
        
        $\mathcal{L}_{\text{reg}} \leftarrow \mathcal{L}_{\text{reg}} + \alpha \mathcal{L}_{\text{smooth}}(\mathbf{h}^{(t)}, \mathbf{W}^{(t)}) + 
            \beta \mathcal{L}_{\text{degree}}(\mathbf{W}^{(t)}) + \gamma \mathcal{L}_{\text{sparse}}(\mathbf{W}^{(t)}))$ using Eqn. \ref{eqn:smoothness}--\ref{eqn:sparse} \tcp{graph regularization loss}
            
        \tcp{step 3. get node embeddings from GNN}
        $\mathbf{z}^{(t)} \leftarrow \text{GNN}(\mathbf{h}^{(t)}, \mathbf{W}^{(t)})$ 
            \tcp{$z^{(t)}$ has shape $(N, 1, D_{hidden})$}
        
    }    
    $\mathbf{Z} \leftarrow \text{concat}(\mathbf{z}^{(1)}, \mathbf{z}^{(2)}, ..., \mathbf{z}^{(n_d)})$ \tcp{concatenate along temporal dim}
    \tcp{step 4. pooling and fully connected layers}
        $\mathbf{Z} \leftarrow \text{graph-pool}\big(\text{temporal-pool}(\mathbf{Z})\big)$
        
        $\hat{\mathbf{y}} \leftarrow \text{Linear}(\mathbf{Z})$
        
        $\mathcal{L}_{\text{pred}} \leftarrow \mathcal{L}_\text{cross-entropy}(\hat{\mathbf{y}}, \mathbf{y})$ \tcp{loss for classification}
   $\mathcal{L} \leftarrow \mathcal{L}_{\text{pred}} + \frac{1}{n_d}\mathcal{L}_{\text{reg}}$ \tcp{total loss}
   
   Back-propagate $\mathcal{L}$ to update model weights $\mathbf{\theta}$ }
   
\end{algorithm2e}

\section{Details of datasets and baselines}
\label{supp:dataset_details}

Table \ref{supp_tab:summary_of_datasets} summarizes the task, sequence length, and number of nodes in each of the three datasets.

\begin{table*}[ht]
\centering
\caption{Summary of datasets.}
\resizebox{\textwidth}{!}{\begin{tabular}{lcccc}
\toprule
Dataset & Biosignal Type & Task                                            & \begin{tabular}[c]{@{}c@{}}Sequence Length \\ (\#Time steps)\end{tabular} & Number of Nodes \\ \midrule
TUSZ    & EEG            & Binary classification of seizure vs non-seizure & 12,000                                                                    & 19              \\
DOD-H   & PSG            & Multiclass classification of five sleep stages  & 7,500                                                                     & 16              \\
ICBEB   & ECG            & Multilabel classification of nine ECG classes   & Varying from 600 to 6,000                                                 & 12             \\ \bottomrule
\end{tabular}}
\label{supp_tab:summary_of_datasets}
\end{table*}

\textbf{Temple University Hospital Seizure Detection Corpus (TUSZ).} We use the publicly available Temporal University Hospital Seizure Detection Corpus (TUSZ) v1.5.2. for seizure detection \citep{Shah2018}. We follow the same experimental setup as in \citet{Tang2022-gnn-ssl}. The TUSZ train set is divided into train and validation splits with distinct patients, and the TUSZ test set is held-out for model evaluation. The following 19 EEG channels are included: FP1, FP2, F3, F4, C3, C4, P3, P4, O1, O2, F7, F8, T3, T4, T5, T6, FZ, CZ, and PZ. Because the EEG signals are sampled at different sampling rate, we resample all the EEG signals to 200 Hz. Following a prior study \citep{Tang2022-gnn-ssl}, we also exclude 5 patients in the TUSZ test set who appear in both the TUSZ train and test sets. The EEG signals are divided into 60-s EEG clips without overlaps, and the task is to predict whether or not an EEG clip contains seizure. Table \ref{supp_tab:tusz_data} shows the number of EEG clips and patients in the train, validation, and test splits.

We compare our model performance to existing models for seizure detection, including (1) LSTM \citep{Hochreiter1997}, a variant of RNN with gating mechanisms; (2) Dense-CNN \citep{Saab2020}, a densely connected CNN specialized in seizure detection; (3) CNN-LSTM \citep{Ahmedt2020-cnnlstm}; and (4) Dist- and Corr-DCRNN without and with self-supervised pre-training \citep{Tang2022-gnn-ssl}. 

\begin{table}[ht]
\centering
\caption{Number of EEG clips and patients in train, validation, and test splits of TUSZ dataset.}
\resizebox{\columnwidth}{!}{\begin{tabular}{lcc} \toprule
               & EEG Clips (\% Seizure) & Patients (\% Seizure) \\ \midrule
Train Set      & 38,613 (9.3\%)         & 530 (34.0\%)          \\
Validation Set & 5,503 (11.4\%)         & 61 (36.1\%)           \\
Test Set       & 8,848 (14.7\%)         & 45 (77.8\%)          \\ \bottomrule
\end{tabular}}
\label{supp_tab:tusz_data}
\end{table}

\textbf{Dreem Open Dataset-Healthy (DOD-H).} We use the publicly available Dreem Open Dataset-Healthy (DOD-H) for sleep staging \citep{Guillot2020-dreem}. The DOD-H dataset consists of overnight PSG sleep recordings from 25 volunteers. The PSG signals are measured from 12 EEG channels, 1 electromyographic (EMG) channel, 2 electrooculography (EOG) channels, and 1 electrocardiogram channel using a Siesta PSG device (Compumedics). All the signals are sampled at 250 Hz. Following the standard AASM Scoring Manual and Recommendations \citep{AASM2018}, each 30-s PSG signal is annotated by a consensus of 5 experienced sleep technologists as one of the five sleep stages: wake, rapid eye movement (REM), non-REM sleep stages N1, N2, and N3. We randomly split the PSG signals by 60/20/20 into train/validation/test splits, where each split has distinct subjects. Table \ref{supp_tab:dodh_data} shows the number of 30-s PSG clips, the number of subjects, and the five sleep stage distributions.

Baseline models include existing sleep staging models that achieved state-of-the-art on DOD-H dataset, SimpleSleepNet \citep{Guillot2020-dreem}, RobustSleepNet \citep{Guillot2021-robustsleepnet}, and DeepSleepNet \citep{Supratak2017-td}, all of which are based on CNNs and/or RNNs. We also include the traditional sequence model LSTM \citep{Hochreiter1997} as a baseline. For fair comparisons between the baselines and \textsc{GraphS4mer}, we trained SimpleSleepNet, RobustSleepNet, and DeepSleepNet using the open sourced code\footnote{SimpleSleepNet: \url{https://github.com/Dreem-Organization/dreem-learning-open}; RobustSleepNet, DeepSleepNet: \url{https://github.com/Dreem-Organization/RobustSleepNet}} and set the temporal context to be one 30-s PSG signal. 

\begin{table}[ht]
\centering
\caption{Number of subjects and 30-s PSG clips in the train, validation, and test splits of DOD-H dataset.}
\resizebox{\columnwidth}{!}{\begin{tabular}{l|c|cccccc}
\toprule
               & \multirow{2}{*}{Subjects} & \multicolumn{6}{c}{30-s PSG Clips}       \\
               &                           & Total  & Wake & N1  & N2   & N3   & REM  \\
               \midrule
Train      & 15                        & 14,823 & 1,839 & 925 & 6,965 & 2,015 & 3,079 \\
Validation & 5                         & 5,114  & 480  & 254 & 2,480 & 990  & 910  \\
Test       & 5                         & 4,725  & 718  & 326 & 2,434 & 509  & 738  \\
\bottomrule
\end{tabular}}
\label{supp_tab:dodh_data}
\end{table}

\textbf{ICBEB ECG Dataset.} We use the publicly available ECG dataset for the 1st China Physiological Signal Challenge held during the International Conference on Biomedical Engineering and Biotechnology (ICBEB) for ECG classification \citep{Liu2018-pu}. We follow the same data split as described in \citet{Strodthoff2021-tx}. Specifically, as the official ICBEB test set is not publicly available, we only use the official ICBEB training set and randomly split it into 10 folds using stratified split, where the first 8 folds are used as the training set, the 9th fold is used as the validation set for hyperparameter tuning, and the 10th fold is used as the held-out test set to report results in this study. In total, there are 6,877 12-lead ECGs ranging between 6-s and 60-s. Following \citet{Strodthoff2021-tx}, the ECGs are downsampled to 100 Hz, resulting in sequence lengths ranging from 600 to 6k time steps. To handle variable length ECGs for training in mini batches, we pad the short sequences with 0s. During training and testing, the padded values are masked out so that they are not seen by the models. In the ICBEB dataset, each ECG record is annotated by up to three reviewers. There are 9 classes in total, including one normal and 8 abnormal classes, and each ECG may be associated with more than one abnormal classes (i.e., multilabel classification). Table \ref{supp_tab:icbeb_data} shows the 9 classes and the number of ECGs in each class in the train/validation/test splits.

We compare \textsc{GraphS4mer} to a wide variety of prior CNNs/RNNs for ECG classification as in \citet{Strodthoff2021-tx}: (1) InceptionTime \citep{Ismail_Fawaz2020-dh}, (2) XResNet1D \citep{He2019-rt}, (3) ResNet1D \citep{Wang2017-sc}, (4) fully convolutional network (FCN) \citep{Wang2017-sc}, (5) LSTM \citep{Hochreiter1997}, (6) bidirectional LSTM \citep{Hochreiter1997}, and (7) WaveletNN \citep{Strodthoff2021-tx}. Note that these baselines take 2.5-s ECG windows as inputs and aggregate the window-wise predictions at test time, whereas \textsc{GraphS4mer} takes the entire ECG signal as input, which allows modeling long-range temporal dependencies in the entire signals. We ran the baseline models using the open source implementations\footnote{\url{https://github.com/helme/ecg_ptbxl_benchmarking}} for three runs with different random seeds.

\begin{table}[ht]
\centering
\caption{Number of ECGs in the train, validation, and test splits of ICBEB dataset used in this study. Note that some ECGs are associated with more than one abnormal classes, and thus the total number of ECGs (last row) is not equal to the sum of the ECGs in the individual classes. \textit{Abbreviations}: AFIB, atrial fibrillation; I-AVB, first-degree atrioventricular block; LBBB, left bundle branch block; RBBB, right bundle branch block; PAC, premature atrial contraction; PVC, premature ventricular contraction; STD, ST-segment depression; STE, ST-segment elevated.}
\begin{tabular}{lccc} \toprule
       & Train & Validation & Test \\ \midrule
Normal & 734   & 92         & 92   \\
AFIB   & 976   & 122        & 123  \\
I-AVB  & 578   & 72         & 72   \\
LBBB  & 189   & 24         & 23   \\
RBBB  & 1,487  & 176        & 194  \\
PAC    & 493   & 61         & 62   \\
PVC    & 560   & 70         & 70   \\
STD    & 695   & 87         & 87   \\
STE    & 176   & 22         & 22   \\ \midrule
Total  & 5,499  & 690        & 688  \\ \bottomrule
\end{tabular}
\label{supp_tab:icbeb_data}
\end{table}

\section{Details of model training procedures and hyperparameters}
\label{supp:model_training}

Model training was accomplished using the AdamW optimizer \citep{loshchilov2018-adamw} in PyTorch on a single NVIDIA A100 GPU. All experiments were run for three runs with different random seeds. Cosine learning rate scheduler with 5-epoch warm start was used \citep{loshchilov2017sgdr}. Model training was early stopped when the validation loss did not decrease for 20 consecutive epochs. We performed the following hyperparameter search on the validation set using an off-the-shelf hyperparameter tuning tool\footnote{\url{https://docs.wandb.ai/guides/sweeps}}: (1) initial learning rate within range [1e-4, 1e-2]; (2) dropout rate in S4 and GNN layers within range [0.1, 0.5]; (3) hidden dimension of S4 and GNN layers within range \{64, 128, 256\}; (4) number of S4 layers within range \{2, 3, 4\}; (5) S4 bidirectionality; (6) number of GNN layers within range \{1, 2\}; (7) graph pooling within range \{mean-pool, max-pool, sum-pool\}; (8) value of $\kappa$ threshold for graph pruning by keeping edges whose weights $> \kappa$ ($\kappa$ within range [0.01, 0.5]); (9) KNN graph with $K \in \{2, 3\}$; (10) weight of KNN graph $\epsilon \in [0.3, 0.6]$; (11) $\alpha$, $\beta$, and $\gamma$ weights in graph regularization within range [0, 1].

\textbf{Model training and hyperparameters for seizure detection on TUSZ dataset.} As there are many more negative samples in the dataset, we undersampled the negative examples in the train set during training. We used binary cross-entropy loss as the loss function. The models were trained for 100 epochs with an initial learning rate of 8e-4. The batch size was 4; dropout rate was 0.1; hidden dimension was 128; number of stacked S4 layers was 4; S4 layers were unidirectional; number of GNN layers was 1; graph pooling was max-pool and temporal pooling was mean-pool; graph pruning was done by setting a threshold of $\kappa = 0.1$, where edges whose edge weights $<=0.1$ were removed; $K=2$ for KNN graph and the KNN graph weight $\epsilon$ was 0.6; $\alpha$, $\beta$, and $\gamma$ weights were all set to 0.05. This results in 265k trainable parameters in \textsc{GraphS4mer}. Best model was picked based on the highest AUROC on the validation set. To obtain binarized predictions, we selected the cutoff threshold by maximizing the F1-score on the validation set.

\textbf{Model training and hyperparameters for sleep staging on DOD-H dataset.} As the DOD-H dataset is highly imbalanced, we undersampled the majority classes in the train set during training. We used cross-entropy loss as the loss function. The models were trained for 100 epochs with an initial learning rate of 1e-3. The batch size was 4; dropout rate was 0.4; hidden dimension was 128; number of stacked S4 layers was 4 and S4 layers were unidirectional; number of GNN layers was 1; graph pooling was sum-pool and temporal pooling was mean-pool; graph pruning was done by setting a threshold of $\kappa = 0.1$; $K=3$ for KNN graph and the weight for KNN graph $\epsilon$ was 0.6; $\alpha$, $\beta$, and $\gamma$ weights were all set to 0.2. This results in 266k trainable parameters in \textsc{GraphS4mer}. Best model was picked based on the highest macro-F1 score on the validation set.

\textbf{Model training and hyperparameters for ECG classification on ICBEB dataset.} As each ECG in the ICBEB dataset may be associated with more than one classes, this task is a multilabel classification task. Therefore, we used binary cross-entropy loss as the loss function. The models were trained for 100 epochs with an initial learning rate of 1e-3. The batch size was 8; dropout rate was 0.1; hidden dimension was 128; number of stacked S4 layers was 4; S4 layers were bidirectional; number of GNN layers was 1; graph pooling was mean-pool and temporal pooling was mean-pool; graph pruning was done by setting a threshold of $\kappa = 0.02$; $K=2$ for KNN graph and the weight for KNN graph $\epsilon$ was 0.6; $\alpha$, $\beta$, and $\gamma$ weights were 1.0, 0.0, and 0.5, respectively. This results in 299k trainable parameters in \textsc{GraphS4mer} for ECG classification. Best model was picked based on the highest macro-AUROC on the validation set. To obtain binarized predictions, we selected the cutoff threshold for each class separately by maximizing the F1-score on the validation set for the respective class.

\section{Evaluation metrics for ECG classification on ICBEB dataset}
\label{supp:icbeb_eval_metrics}
In binary classification, $F_{\beta}$ score uses a positive real factor $\beta$ to weigh recall $\beta$ times as important as precision:
\begin{align*}
F_{\beta} = (1 + \beta^2) \times \frac{\text{Precision} \times \text{Recall}}{(\beta^2 \times \text{Precision}) + \text{Recall}}
\end{align*}

Specifically, F1-score (i.e., $\beta = 1$) is the harmonic mean of precision and recall: 
\begin{align*}
F1 = \frac{2 \times \text{Precision} \times \text{Recall}}{\text{Precision} + \text{Recall}}
\end{align*}

F2-score considers recall more important than precision:
\begin{align*}
F2 = \frac{(1+2^2) \times \text{Precision} \times \text{Recall}}{(2^2 \times \text{Precision}) + \text{Recall}}
\end{align*}

In contrast, $G_{\beta}$ measure is defined as:
\begin{align*}
    G_{\beta} = \frac{TP}{(TP + FP + \beta \times FN)}
\end{align*}
where TP, FP, and FN are the number of true positives, false positives, and false negatives, respectively. $G2$-score is calculated by setting $\beta=2$ in the above equation.

For multilabel/multiclass classification, macro-averaged scores are computed by taking the mean of the individual scores in the individual classes. For ECG classification experiments, we report macro-F1, macro-F2, macro-G2, and macro-AUROC in Table \ref{tab:icbeb_results}.

\section{Model performance on individual ECG classes}
\label{supp:ecg_indiv_classes}

Table \ref{supp_tab:ecg_indiv_class} shows \textsc{GraphS4mer} and baseline model performance for individual ECG classes. \textsc{GraphS4mer} outperforms the baselines on 6 out of 9 classes.

\begin{table*}[ht]
\centering
\caption{Model performance (AUROC) for individual ECG classes. Best and second best results are \textbf{bolded} and \underline{underlined}, respectively.}
\resizebox{\textwidth}{!}{\begin{tabular}{lccccccccc}
\toprule
Model         & Normal       & AFIB         & I-AVB        & LBBB         & RBBB         & PAC          & PVC          & STD          & STE          \\ \midrule
InceptionTime & \underline{0.962 $\pm$ 0.001} & 0.973 $\pm$ 0.004 & 0.992 $\pm$ 0.001 & 0.972 $\pm$ 0.018 & \underline{0.990 $\pm$ 0.003} & 0.913 $\pm$ 0.022 & 0.970 $\pm$ 0.004 & \underline{0.956 $\pm$ 0.004} & 0.946 $\pm$ 0.025 \\
XResNet1D     & \underline{0.962 $\pm$ 0.000} & \textbf{0.977 $\pm$ 0.001} & 0.985 $\pm$ 0.002 & \underline{0.973 $\pm$ 0.004} & \textbf{0.991 $\pm$ 0.001} & \underline{0.955 $\pm$ 0.003} & \underline{0.978 $\pm$ 0.002} & 0.948 $\pm$ 0.003 & \textbf{0.966 $\pm$ 0.012} \\
ResNet1D      & 0.960 $\pm$ 0.003 & \underline{0.976 $\pm$ 0.002} & \underline{0.993 $\pm$ 0.002} & 0.963 $\pm$ 0.019 & 0.989 $\pm$ 0.001 & 0.905 $\pm$ 0.007 & 0.975 $\pm$ 0.004 & 0.951 $\pm$ 0.005 & \underline{0.957 $\pm$ 0.014} \\
FCN           & 0.955 $\pm$ 0.002 & 0.972 $\pm$ 0.002 & 0.986 $\pm$ 0.002 & 0.956 $\pm$ 0.008 & \underline{0.990 $\pm$ 0.000} & 0.834 $\pm$ 0.013 & 0.971 $\pm$ 0.002 & 0.953 $\pm$ 0.000 & 0.929 $\pm$ 0.013 \\
LSTM          & 0.903 $\pm$ 0.007 & 0.973 $\pm$ 0.004 & 0.984 $\pm$ 0.003 & 0.934 $\pm$ 0.009 & 0.976 $\pm$ 0.004 & 0.886 $\pm$ 0.014 & 0.942 $\pm$ 0.008 & 0.954 $\pm$ 0.007 & 0.929 $\pm$ 0.040 \\
Bidir-LSTM    & 0.917 $\pm$ 0.003 & 0.975 $\pm$ 0.005 & 0.991 $\pm$ 0.001 & 0.937 $\pm$ 0.012 & 0.984 $\pm$ 0.003 & 0.899 $\pm$ 0.009 & 0.945 $\pm$ 0.002 & 0.946 $\pm$ 0.003 & 0.915 $\pm$ 0.035 \\
WaveletNN     & 0.945 $\pm$ 0.002 & 0.921 $\pm$ 0.006 & 0.901 $\pm$ 0.002 & 0.965 $\pm$ 0.005 & 0.947 $\pm$ 0.002 & 0.737 $\pm$ 0.005 & 0.928 $\pm$ 0.004 & 0.922 $\pm$ 0.003 & 0.932 $\pm$ 0.015 \\ \midrule
\textsc{GraphS4mer} (ours)    & \textbf{0.989 $\pm$ 0.003} & 0.970 $\pm$ 0.002 & \textbf{0.996 $\pm$ 0.001} & \textbf{0.978 $\pm$ 0.006} & 0.986 $\pm$ 0.001 & \textbf{0.965 $\pm$ 0.004} & \textbf{0.983 $\pm$ 0.002} & \textbf{0.971 $\pm$ 0.005} & 0.952 $\pm$ 0.007
\\ \bottomrule
\end{tabular}}
\label{supp_tab:ecg_indiv_class}
\end{table*}

\begin{table*}[ht]
\centering
\caption{Additional evaluation metrics for ablation experiments. Best and second best results are \textbf{bolded} and \underline{underlined}, respectively.}
\resizebox{\textwidth}{!}{\begin{tabular}{l|cccc|c|ccc}
\toprule
                   & \multicolumn{4}{c|}{TUSZ}                                  & DOD-H        & \multicolumn{3}{c}{ICBEB}                      \\
Model              & F1-Score     & AUPRC        & Sensitivity  & Specificity  & Kappa        & Macro F2-Score & Macro G2-Score & Macro AUROC  \\ \midrule
S4                 & 0.491 $\pm$ 0.033 & 0.488 $\pm$ 0.017 & 0.441 $\pm$ 0.077 & \textbf{0.941 $\pm$ 0.020} & 0.757 $\pm$ 0.012 & 0.789 $\pm$ 0.004   & 0.577 $\pm$ 0.006   & 0.966 $\pm$ 0.004 \\
\textsc{GraphS4mer} w/o S4  & 0.373 $\pm$ 0.100 & 0.302 $\pm$ 0.117 & 0.537 $\pm$ 0.133 & 0.765 $\pm$ 0.065 & 0.597 $\pm$ 0.064 & 0.313 $\pm$ 0.006   & 0.107 $\pm$ 0.002   & 0.529 $\pm$ 0.028 \\
GSL-GNN-S4         & 0.637 $\pm$ 0.010 & 0.668 $\pm$ 0.019 & 0.629 $\pm$ 0.053 & \textbf{0.941 $\pm$ 0.014} & \underline{0.784 $\pm$ 0.014} & 0.782 $\pm$ 0.016   & 0.564 $\pm$ 0.019   & 0.966 $\pm$ 0.001 \\
\textsc{GraphS4mer} w/o GSL & \underline{0.672 $\pm$ 0.029} & \underline{0.672 $\pm$ 0.029} & \underline{0.708 $\pm$ 0.020} & 0.931 $\pm$ 0.019 & 0.734 $\pm$ 0.010 & \underline{0.794 $\pm$ 0.011}  & \underline{0.588 $\pm$ 0.015}   & \underline{0.970 $\pm$ 0.007} \\ \midrule
\textsc{GraphS4mer} (ours)  & \textbf{0.680 $\pm$ 0.012} & \textbf{0.723 $\pm$ 0.023} & \textbf{0.718 $\pm$ 0.041} & \underline{0.933 $\pm$ 0.009} & \textbf{0.802 $\pm$ 0.014} & \textbf{0.804 $\pm$ 0.004}   & \textbf{0.609 $\pm$ 0.005}   & \textbf{0.977 $\pm$ 0.001} \\ \bottomrule
\end{tabular}}
\label{supp_tab:ablation_more_metrics}
\end{table*}

\section{Additional evaluation metrics for ablation experiments}
\label{supp:ablation_more_metrics}
Table \ref{supp_tab:ablation_more_metrics} shows additional evaluation metrics for ablation experiments on TUSZ, DOD-H, and ICBEB datasets.

\section{Effect of temporal resolution}
\label{supp:effect_of_r}
\begin{figure}[ht!]
    \centering
    \includegraphics[width=0.8\columnwidth]{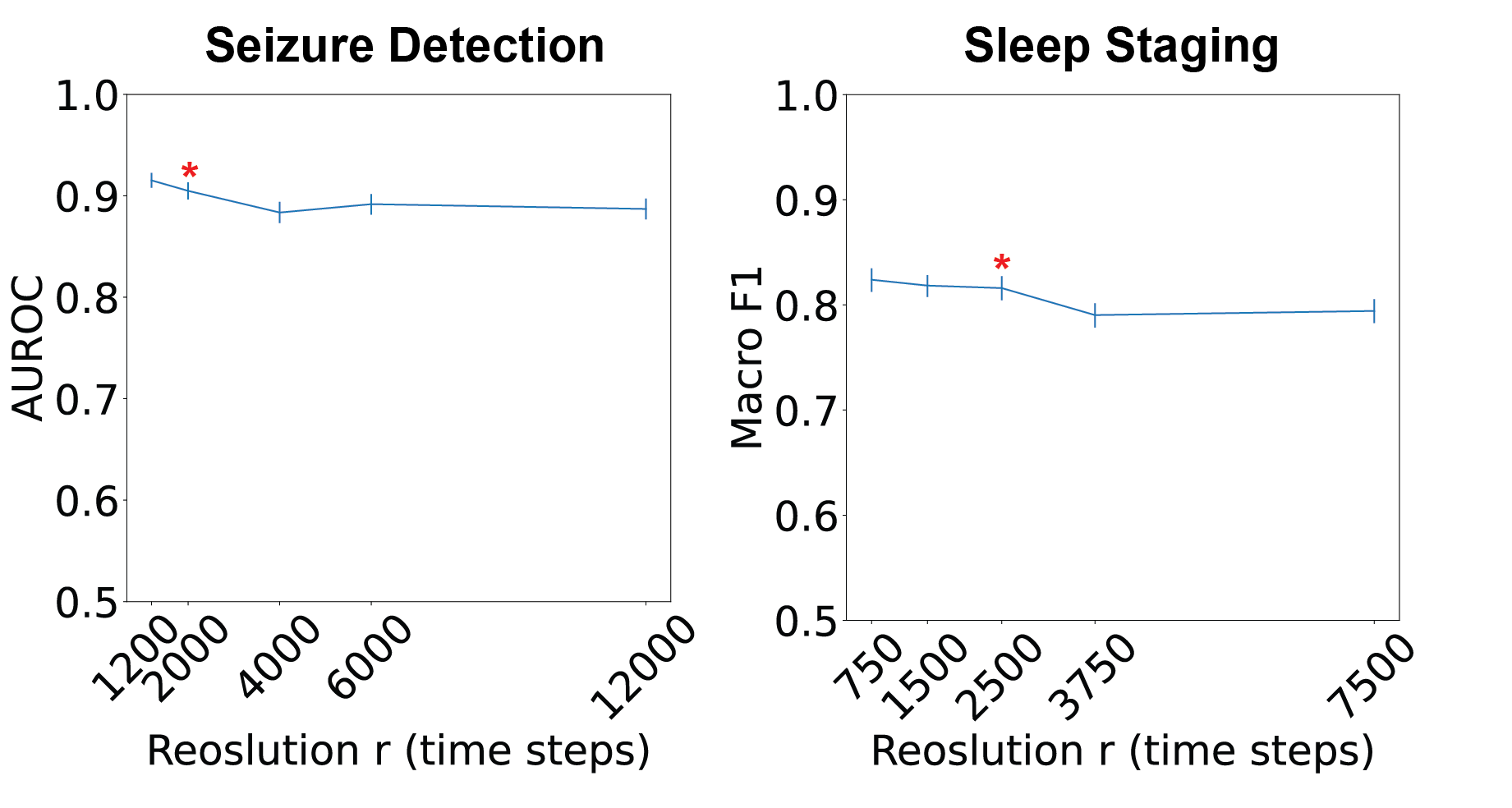}
    \caption{Model performance versus temporal resolution using the run with median performance. For convenience, we assume that temporal resolution is chosen so that the sequence length is divisible by the resolution. Asterisk indicates the temporal resolution used to report results for \textsc{GraphS4mer} in Tables \ref{tab:tuh_results}--\ref{tab:dodh_results} and Table \ref{tab:ablation}. Error bars indicate 95\% confidence intervals obtained using bootstrapping with 5,000 replicates with replacement.}
    \label{supp_fig:effect_of_r}
\end{figure}

To examine the effect of temporal resolution $r$ on model performance, we show the performance versus different values of $r$ for seizure detection and sleep staging in Figure \ref{supp_fig:effect_of_r}. We observe that smaller value of $r$ tends to result in higher performance, which suggests that capturing dynamically varying graph structures is useful for these tasks.

\section{Effect of KNN graph weight and graph regularization}
\label{supp:effect_of_knn_graph_reg}

\begin{figure}[ht!]
    \centering
    \includegraphics[width=\columnwidth]{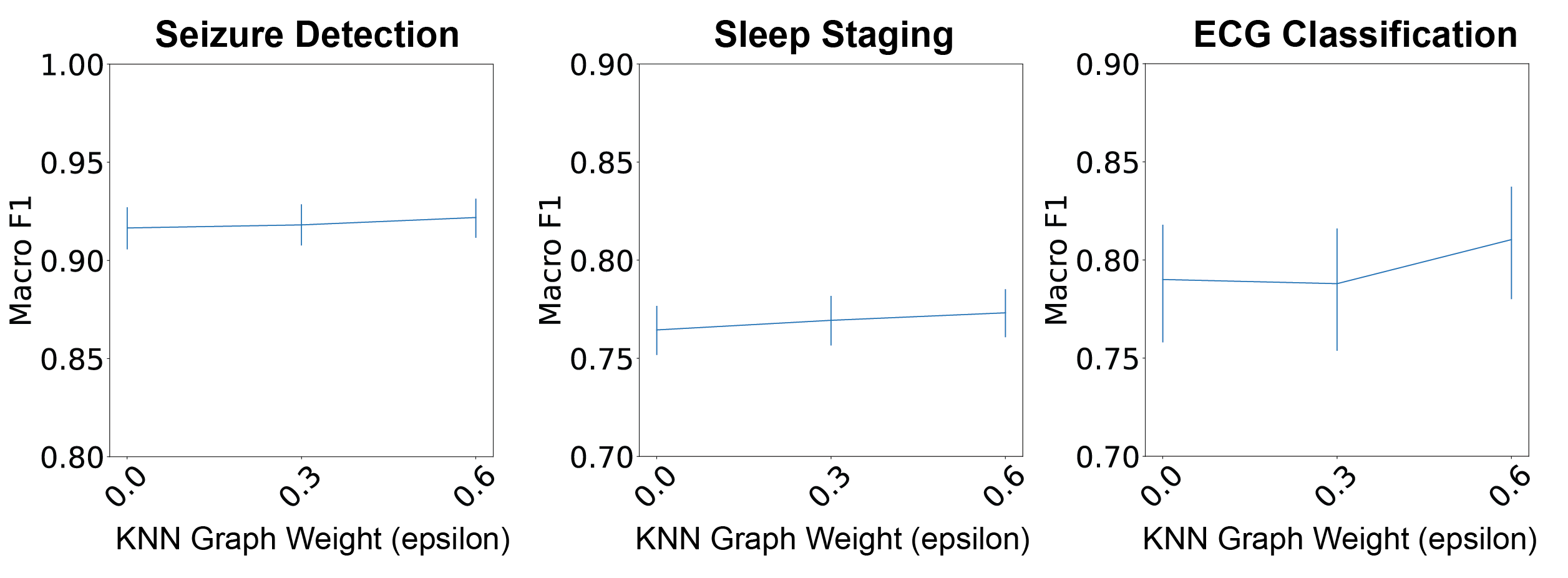}
    \caption{Model performance on the validation set versus KNN graph weight ($\epsilon$ in Equation \ref{eqn:add_knn_graph}) using the run with median performance. Error bars indicate 95\% confidence intervals obtained using bootstrapping with 5,000 replicates with replacement.}
    \label{supp_fig:effect_of_knn}
\end{figure}

Figure \ref{supp_fig:effect_of_knn} shows the validation set performance for \textsc{GraphS4mer} with respect to different values of KNN graph weight (i.e., $\epsilon$ in Equation \ref{eqn:add_knn_graph}). A larger KNN graph weight results in marginally better model performance.

\begin{figure}[ht!]
    \centering
    \includegraphics[width=\columnwidth]{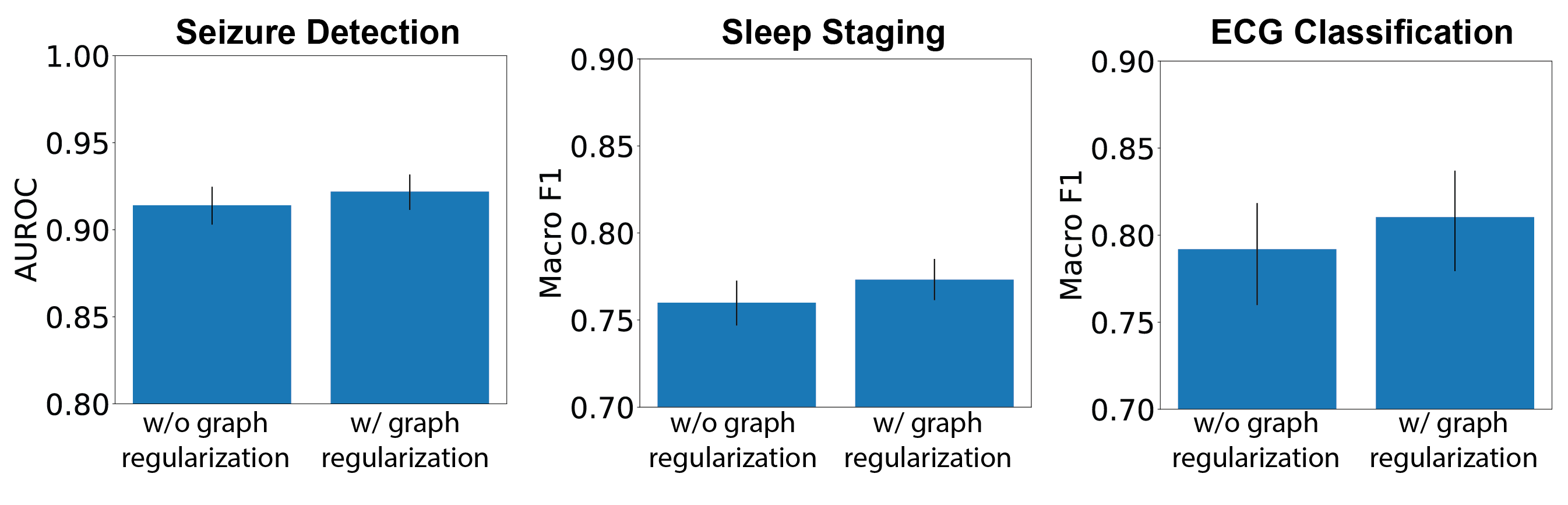}
    \caption{Model performance on the validation set for \textsc{GraphS4mer} without and with graph regularization loss using the run with median performance. Error bars indicate 95\% confidence intervals obtained using bootstrapping with 5,000 replicates with replacement.}
    \label{supp_fig:effect_of_graph_reg}
\end{figure}

Figure \ref{supp_fig:effect_of_graph_reg} shows the validation set performance for \textsc{GraphS4mer} without and with graph regularization loss. Graph regularization results in marginally better model performance.

\section{Computational cost of GSL}
\label{supp:comp_cost}

\begin{table*}[ht!]
\centering
\caption{Number of additional parameters and multiply–accumulate operations (MACs) with respect to \textsc{GraphS4mer} w/o GSL (with predefined graphs). The additional parameters and MACs come from the self-attention layer in GSL, and the MACs increase as the time interval $r$ decreases. $r$ indicates the time interval (resolution) in the GSL layer, $T$ indicates the sequence length (dependent on dataset), and $n_d$ indicates the number of dynamic graphs. For convenience, we assume that the resolution $r$ is chosen so that the sequence length $T$ is divisible by $r$. ``-" indicates that the sequence length is not divisible by $r$. ``NA" indicates not applicable, as we set $r$ to be the actual sequence lengths of ECGs in the ICBEB dataset due to variable sequence lengths. Asterisk indicates the resolution used to report results for \textsc{GraphS4mer} in Tables \ref{tab:tuh_results}--\ref{tab:ablation}.}
\resizebox{\textwidth}{!}{\begin{tabular}{l|cc|cc|cc}
\toprule
\multirow{2}{*}{Model}       & \multicolumn{2}{c|}{TUSZ ($T=12,000$)} & \multicolumn{2}{c|}{DOD-H ($T=7,500$)} & \multicolumn{2}{c}{ICBEB (varying $T$)} \\
                             & \#Parameters  & MACs     & \#Parameters  & MACs      & \#Parameters   & MACs     \\ \midrule
\textsc{GraphS4mer} w/o GSL           & ref           & ref      & ref           & ref       & ref            & ref      \\ \midrule
\textsc{GraphS4mer} ($r=T, n_d=1$)     & 32.768K       & 19.923M  & 32.768K       & 16.777M   & 32.768K$^*$        & 12.583M$^*$  \\
\textsc{GraphS4mer} ($r=T/2, n_d=2$)   & 32.768K       & 39.846M  & 32.768K       & 33.554M   & NA             & NA       \\
\textsc{GraphS4mer} ($r=T/3, n_d=3$)   & 32.768K       & 59.769M  & 32.768K$^*$       & 50.332M$^*$   & NA             & NA       \\
\textsc{GraphS4mer} ($r=T/4, n_d=4$)   & 32.768K       & 79.692M  & 32.768K       & 67.109M   & NA             & NA       \\
\textsc{GraphS4mer} ($r=T/5, n_d=5$)   & 32.768K       & 99.615M  & 32.768K       & 83.886M   & NA             & NA       \\
\textsc{GraphS4mer} ($r=T/6, n_d=6$)   & 32.768K$^*$       & 119.538M$^*$ & 32.768K       & 100.663M  & NA             & NA       \\
\textsc{GraphS4mer} ($r=T/8, n_d=8$)   & 32.768K       & 159.384M & -       & -         & NA             & NA       \\
\textsc{GraphS4mer} ($r=T/10, n_d=10$) & 32.768K       & 199.229M & 32.768K       & 167.772M  & NA             & NA       \\ \bottomrule
\end{tabular}}
\label{supp_tab:comp_cost}
\end{table*}

In Table \ref{supp_tab:comp_cost}, we compare the additional number of parameters and multiply-accumulate operations (MACs) of \textsc{GraphS4mer} with respect to \textsc{GraphS4mer} without GSL (with predefined graphs), as well as computational costs of \textsc{GraphS4mer} with different values of resolutions $r$ in the GSL layer (see Section \ref{section:gsl}). The GSL layer (i.e., self-attention) adds additional 33k parameters (approximately 12\% of total parameters). Moreover, as the time interval $r$ decreases (i.e., number of dynamic graphs increases), the number of MACs increases, whereas the number of parameters remains the same. This is expected given that the same GSL layer is applied multiple times to obtain dynamic graphs. Specifically, the additional MACs are proportional to the number of dynamic graphs in the GSL layer.

\section{Visualization of PSG graphs for five sleep stages}
\label{supp:sleep_adj_all}

\begin{figure*}[h!]
    \centering
    \includegraphics[width=\textwidth]{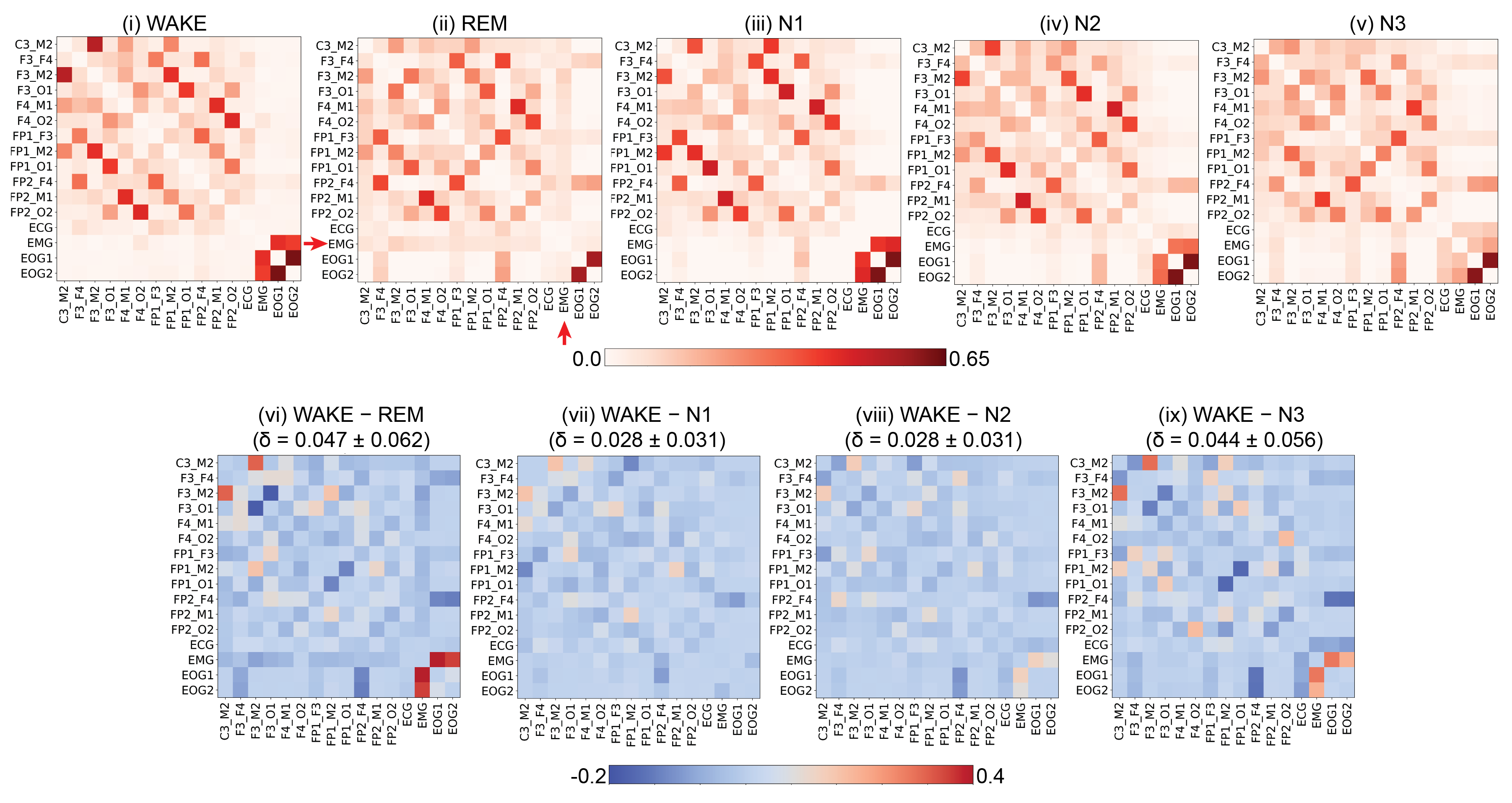}
    \caption{\textbf{(i)--(v)} Mean adjacency matrices for PSG signals for five sleep stages in correctly predicted test samples. \textbf{(vi)--(ix)} Difference between non-WAKE stages and WAKE. $\delta$ indicates the mean and standard deviation of the absolute values of the differences between non-WAKE stages and WAKE. Red arrow indicates EMG channel in REM stage that has weak connection to all other channels. Self-edges (i.e., diagonal) are not shown here.}
    \label{supp_fig:sleep_adj_all}
\end{figure*}

Figure \ref{supp_fig:sleep_adj_all} shows mean adjacency matrices for PSG signals in correctly predicted test samples grouped by five sleep stages. We observe that N3 (Figure \ref{supp_fig:sleep_adj_all}ix) differs from wake stage more than N1 (Figure \ref{supp_fig:sleep_adj_all}vii) (also see Figure \ref{fig:viz_eeg_sleep_adj}). Moreover, in REM stage (Figure \ref{supp_fig:sleep_adj_all}ii), the EMG channel has very weak connection to all other channels (red arrow in Figure \ref{supp_fig:sleep_adj_all}ii), which is expected given that one experiences muscle paralysis in REM stage.

\section{Visualization of example ECG signals and ECG graphs}
\label{supp:ecg_adj_all}

\begin{figure*}[h!]
    \centering
    \includegraphics[width=\textwidth]{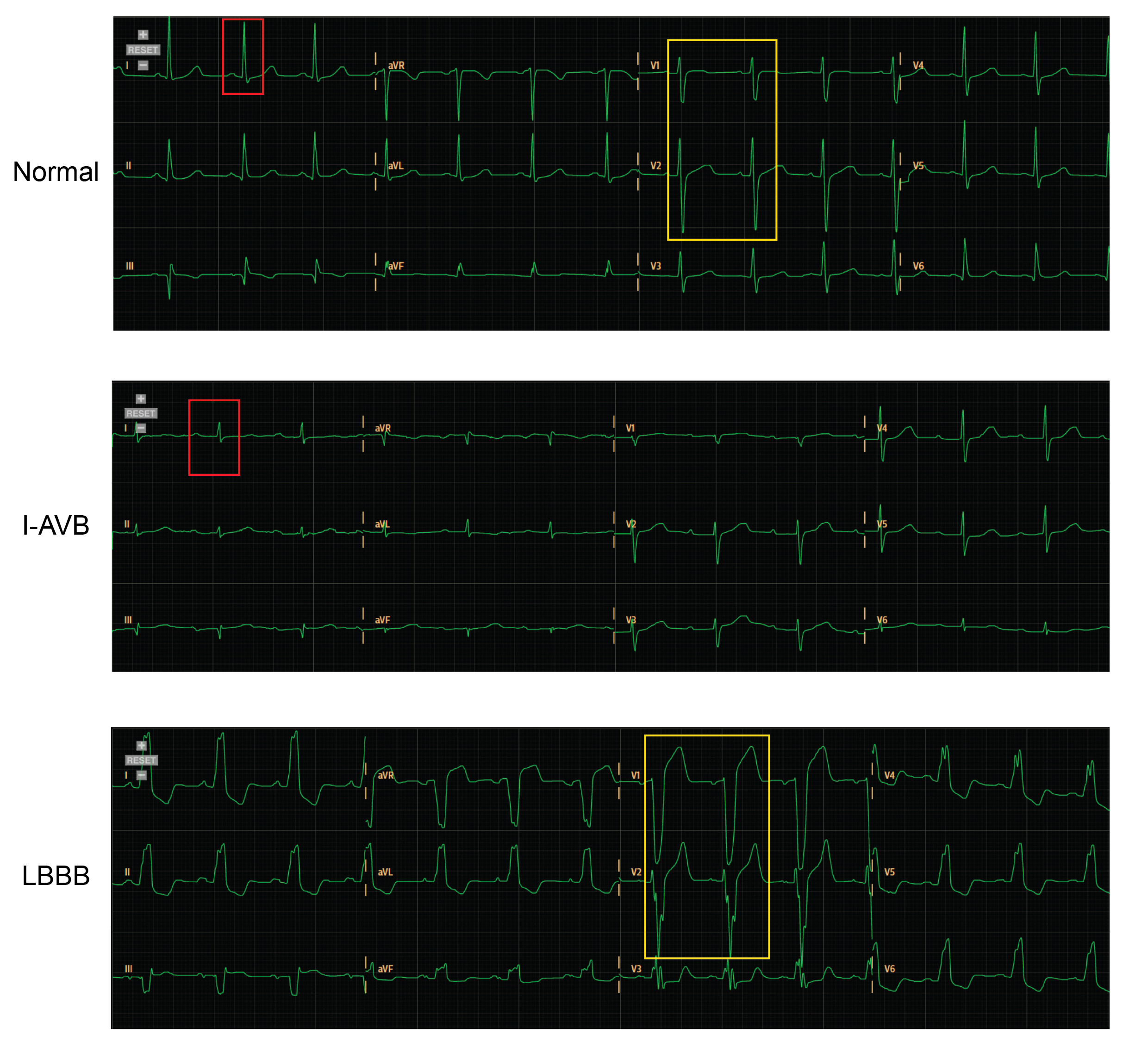}
    \caption{Example normal ECG, ECG with first-degree atrioventricular block (I-AVB), and ECG with left bundle branch block (LBBB). Red boxes demonstrate the difference in PR interval between I-AVB and normal ECG. Yellow boxes demonstrate the pronounced morphology difference in the QRS complex between LBBB and normal ECG.}
    \label{supp_fig:ecg_examples}
\end{figure*}

Figure \ref{supp_fig:ecg_examples} shows example normal ECG, ECG with first-degree atrioventricular block (I-AVB), and ECG with left bundle branch block (LBBB).

\begin{figure*}[h!]
    \centering
    \includegraphics[width=\textwidth]{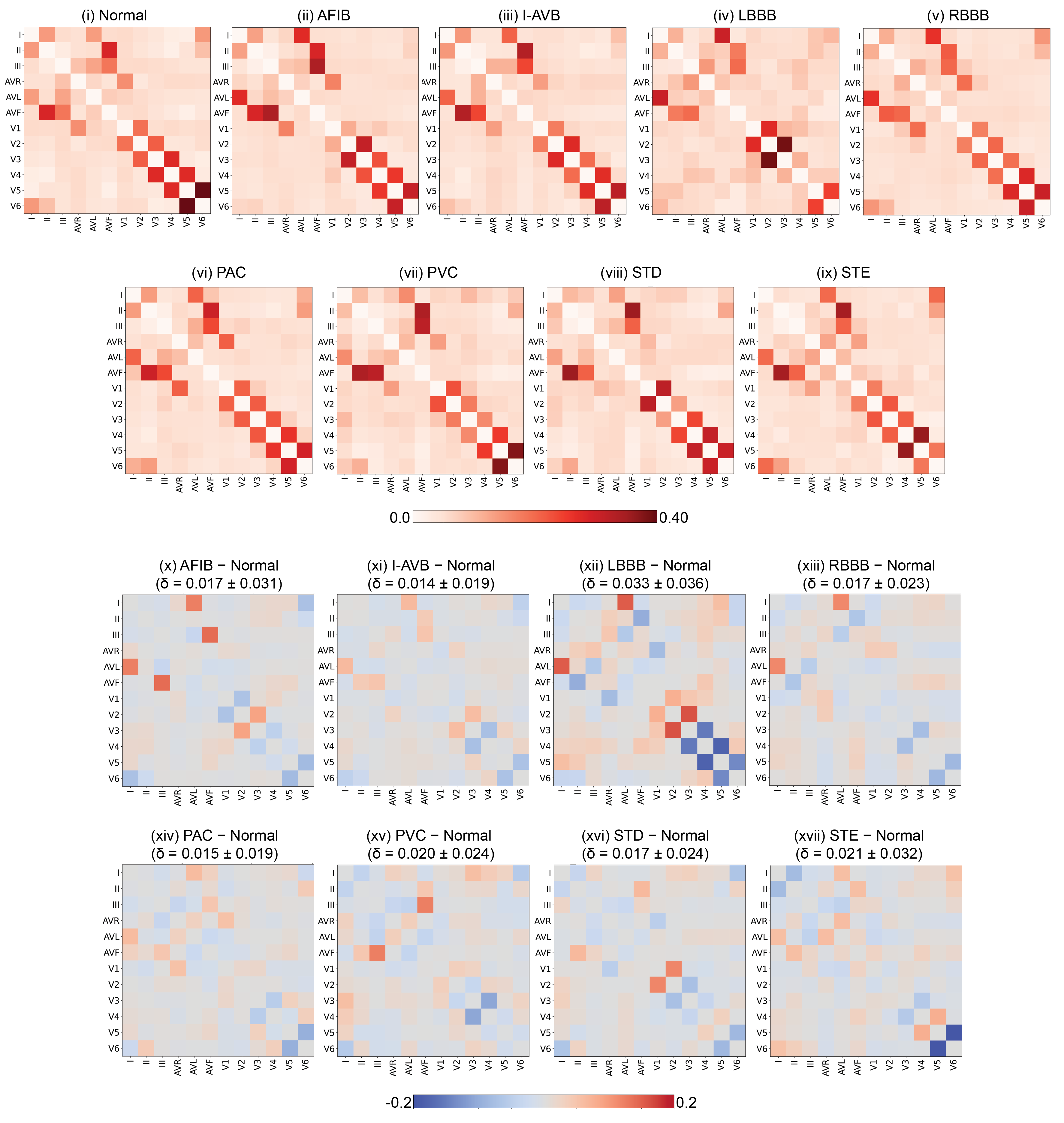}
    \caption{\textbf{(i)--(ix)} Mean adjacency matrices for ECG signals for nine ECG classes in correctly predicted test samples. \textbf{(x)--(xvii)} Difference between adjacency matrices of abnormal ECG classes and normal class. $\delta$ indicates the mean and standard deviation of the absolute values of the differences between the abnormal class and the normal class. Self-edges (i.e., diagonal) are not shown here. \textit{Abbreviations}: AFIB, atrial fibrillation; I-AVB, first-degree atrioventricular block; LBBB, left bundle branch block; RBBB, right bundle branch block; PAC, premature atrial contraction; PVC, premature ventricular contraction; STD, ST-segment depression; STE, ST-segment elevated.}
    \label{supp_fig:ecg_adj_all}
\end{figure*}

Figure \ref{supp_fig:ecg_adj_all} shows mean adjacency matrices for ECG signals in correctly predicted test samples, grouped by nine ECG classes. $\delta$ indicates the mean and standard deviation of the absolute values of the differences between the abnormal class and the normal class. We observe that the magnitude of difference between I-AVB and normal ECG is the smallest (i.e., smallest $\delta$), whereas the magnitude of difference between LBBB and normal ECG is the largest (i.e., largest $\delta$).

\end{document}